\documentclass[runningheads]{llncs}
\usepackage{graphicx}

\usepackage{tikz}
\usepackage{comment}
\usepackage{amsmath,amssymb} 
\usepackage{color}

\usepackage{amsthm}
\usepackage{subcaption}
\usepackage{booktabs}
\usepackage{multirow}
\usepackage{pbox}
\usepackage[utf8]{inputenc}
\usepackage{mathtools}
\usepackage[export]{adjustbox} 
\usepackage{longtable}
\usepackage{rotating}
\usepackage{adjustbox}
\usepackage{enumitem}

\usepackage{import}
\usepackage{extarrows}
\usepackage{xparse} 
\usepackage{calc}   
\usepackage{colortbl}
\usepackage{xfrac}
\usepackage{tablefootnote}
\usepackage{tabularx}
\usepackage{standalone}
\definecolor{rowblue}{RGB}{220,230,240}
\definecolor{Orange}{RGB}{255,204,153}
\definecolor{Blue}{RGB}{153,204,255}

\usepackage{amsmath}

\DeclareMathOperator*{\argmin}{arg\,min}

\usepackage[linesnumbered,boxed,noend]{algorithm2e}
\SetKwInput{KwInput}{Input}
\SetKwInput{KwOutput}{Output}

\makeatletter
\define@key{Gin}{expandoptions}{%
  \begingroup
  \edef\x{\endgroup
    \noexpand\setkeys{Gin}{#1}%
  }\x
}
\makeatother

\def\p{\mathbf{p}}%
\def\x{\mathbf{x}}%
\usepackage{pifont}

\usepackage{url}            %
\usepackage[pagebackref=true,breaklinks,colorlinks,bookmarks=false]{hyperref}

\usepackage{mathrsfs}

\usepackage{microtype}

\usepackage{wrapfig}

\newcommand{\mbw}{\mathbf{w}}
\newcommand{\mbC}{\mathbf{C}}
\newcommand{\mbc}{\mathbf{c}}
\newcommand{\mbA}{\mathbf{A}}

\newcommand{\mbP}{\mathbf{P}}
\newcommand{\mbQ}{\mathbf{Q}}

\newcommand{\mbx}{\mathbf{x}}
\newcommand{\mbs}{\mathbf{s}}
\newcommand{\mbI}{\mathbf{I}}

\usepackage{tocloft}

\cftsetindents{section}{0em}{2em}
\cftsetindents{subsection}{0em}{2em}

\renewcommand{\orcidID}[1]{\href{https://orcid.org/#1}{~\includegraphics[width=9pt]{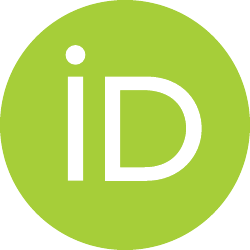}}}

\begin{document}

\pagestyle{headings}
\mainmatter
\def\ECCVSubNumber{1732}  

\title{Generating Handwriting via \\Decoupled Style Descriptors}

\titlerunning{Generating Handwriting via Decoupled Style Descriptors}
\authorrunning{A.~Kotani, S.~Tellex, and J.~Tompkin.}
\author{Atsunobu Kotani\orcidID{0000-0001-6117-6630} \and
Stefanie Tellex\orcidID{0000-0002-2905-4075} \and
James Tompkin\orcidID{0000-0003-2218-2899}}
\institute{Brown University}

\maketitle


\begin{abstract}
Representing a space of handwriting stroke styles includes the challenge of representing both the style of each character and the overall style of the human writer.
%
Existing VRNN approaches to representing handwriting often do not distinguish between these different style components, which can reduce model capability.
%
Instead, we introduce the Decoupled Style Descriptor (DSD) model for handwriting, which factors both character- and writer-level styles and allows our model to represent an overall greater space of styles.
%
This approach also increases flexibility: given a few examples, we can generate handwriting in new writer styles, and also now generate handwriting of new characters across writer styles. 
In experiments, our generated results were preferred over a state of the art baseline method $88\%$ of the time, and in a writer identification task on $20$ held-out writers, our DSDs achieved $89.38\%$ accuracy from a single sample word.
%
Overall, DSDs allows us to improve both the quality and flexibility over existing handwriting stroke generation approaches.
\end{abstract}

\section{Introduction}

Producing computational models of handwriting is a deeply \emph{human} and \emph{personal} topic---most people can write, and each writer has a unique style to their script.
%
%
Capturing these styles flexibly and accurately is important as it determines the space of descriptive expression of the model; in turn, these models define the usefulness of our recognition and generation applications.
For deep-learning-based models, our concern is how to architecture the neural network such that we can represent the underlying stroke characteristics of the styles of writing.

Challenges in handwriting representation include reproducing fine detail, generating unseen characters, enabling style interpolation and transfer, and using human-labeled training data efficiently. Across these, one foundational problem is how to succinctly represent both the style variation of each character and the overall style of the human writer---to capture both the variation within an `h' letterform and the overall consistency with other letterform for each writer.

As handwriting strokes can be modeled as a sequence of points over time, supervised deep learning methods to handwriting representation can use recurrent neural networks (RNNs) \cite{graves2013generating,Aksan:2018:DeepWriting}. 
This allows consistent capture of style features that are distant in time and, with the use of variational RNNs (VRNNs), allows the diverse generation of handwriting by drawing from modeled distributions. 
However, the approach of treating handwriting style as a `unified' property of a sequence can limit the representation of both character- and writer-level features. 
This includes specific character details being averaged out to maintain overall writer style, and an reduced representation space of writing styles.

Instead, we explicitly represent 1) writer-, 2) character- and 3) writer-character-level style variations within an RNN model.
We introduce a method of Decoupled Style Descriptors (DSD) that models style variations such that character style can still depend on writer style. 
Given a database of handwriting strokes as timestamped sequences of points with character string labels \cite{marti2002iam}, we learn a representation that encodes three key factors: %
writer-independent character representations ($\mbC_h$ for character \textit{h}, $\mbC_{his}$ for the word \textit{his}), %
writer-dependent character-string style descriptors ($\mbw_h$ for character \textit{h}, $\mbw_{his}$ for the word \textit{his}), %
and writer-dependent global style descriptors ($\mbw$ per writer).
%
%
This allows new sequence generation for existing writers (via new $\mbw_{she}$), new writer generation via style transfer and interpolation (via new $\mbw$), and new character generation in the style of existing writers (via new $\mbC_\textit{2}$, from only a few samples of character \textit{2} from \emph{any} writer).
Further, our method helps to improve generation quality as more samples are provided for projection, rather than tending towards average letterforms in existing VRNN models.

In a qualitative user study, our model's generations were preferred $88\%$ of the time over an existing baseline \cite{Aksan:2018:DeepWriting}. For writer classification tasks on a held-out $20$-way test, our model achieves accuracy of $89.38\%$ from a single word sample, and $99.70\%$ from 50 word-level samples. %
In summary, we contribute:
\begin{itemize}[topsep=0.5pt,partopsep=0.5pt] 
    \item Decoupled Style Descriptors as a way to represent latent style information;
    \item An architecture with DSDs to model handwriting, with demonstration applications in generation, recognition, and new character adaptation; and
    \item A new database---BRUSH (BRown University Stylus Handwriting)---of handwritten digital strokes in the Latin alphabet, which includes $170$ writers, $86$ characters, $488$ common words written by all writers, and $3668$ rarer words written across writers.
\end{itemize}
Our dataset, code, and model will be open source at \href{http://dsd.cs.brown.edu}{http://dsd.cs.brown.edu}.

\section{Related Work}
\label{sec:relatedwork}

\begin{figure}[t]
	\centering
	\includegraphics[width=0.9\linewidth]{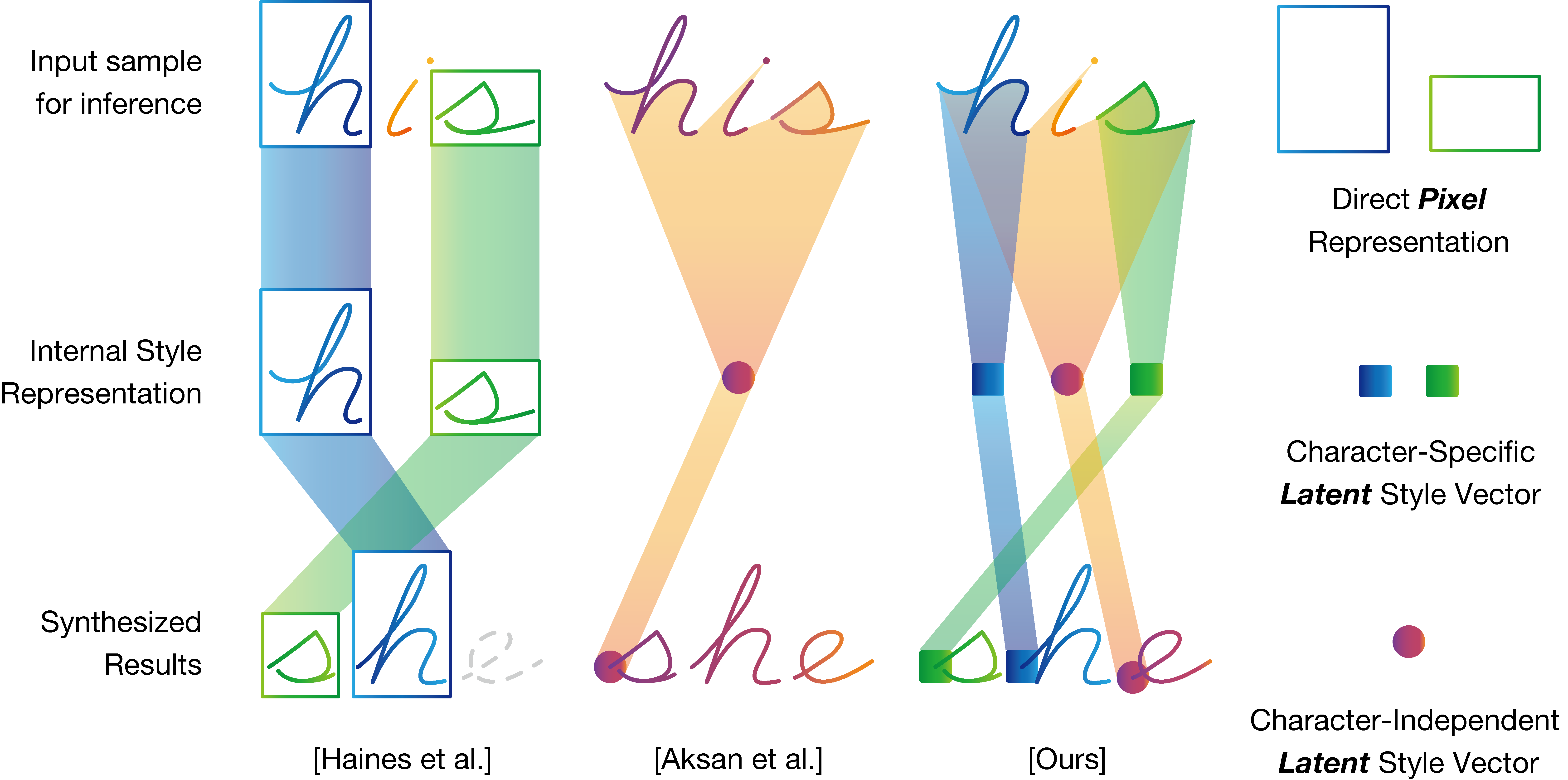}
    \caption{Illustrating synthesis approaches. Given test sample \textit{his} for reference, we wish to generate \textit{she} in the same style.
    Left: Pixels of \textit{h} and \textit{s} are copied from input with a slight modification \cite{mytext}; however, this fails to synthesize \textit{e} as it is missing in the reference.
    Middle: A global latent writer style is inferred from \textit{his} and used as the initial state for LSTM generation \cite{Aksan:2018:DeepWriting}. 
    Right: Our approach infers both character and writer style vectors to improve quality and flexibility.
    }
    \label{fig:his_she}
\end{figure}

Handwriting modeling methods either handle images, which capture writing appearance, or handle the underlying strokes collected via digital pens. 
Each may be online, where observation happens along with writing, or offline. 
Offline methods support historical document analysis, but cannot capture the motion of writing. We consider an online stroke-based approach, which avoids the stroke extraction problem and allows us to focus on modelling style variation. Work also exists in the separate problem of typeface generation~\cite{campbell2014learning,doi:10.1111/cgf.13540,Lopes_2019_ICCV,hu2001parameterizable,zongker2000example}.

\paragraph{General style transfer methods.} Current state-of-the-art style transfer works use a part of the encoded reference sample as a style component, e.g., the output of a CNN encoder for 2D images~\cite{Huang_2017_ICCV,kotovenko2019content}, or the last output of an LSTM for speech audio~\cite{qian2019autovc}. 
These can be mixed to allocate parts of a conditioning style vector to disentangled variation \cite{Hu_2018_CVPR}.
Common style representations often cannot capture small details, with neural networks implicitly filtering out this content information, because the representations fail to structurally decouple style from content in the style reference source. Other approaches~\cite{jia2016dynamic,wang2017zm} tackle this problem by making neural networks predict parameters for a transformation model (an idea that originates from neuroevolution~\cite{stanley2009hypercube,ha2016hypernetworks}); our $\mbC$ prediction is related.

\paragraph{Recent image-based offline methods.}
Haines et al.~produced a system to synthesize new sentences in a specific style inferred from source images \cite{mytext}. 
Some human intervention is needed during character segmentation, and the model can only recreate characters that were in the source images. 
Alonso et al.~addressed the labeling issue with a GAN-based approach~\cite{goodfellow2014generative,Alonso2019AdversarialGO}; however, their model presents an image quality trade-off and struggles to generate new characters.
There are also studies on typeface generation from few reference data~\cite{azadi2018multi,suveeranont2010example}:
Baluja generates typefaces for Latin alphabets~\cite{baluja2017learning},
and Lian et al.~for Chinese~\cite{lian2018easyfont}.
Our method does not capture writing implement appearance, but does provides underlying stroke generation and synthesizes new characters from few examples.

\paragraph{Stroke-based online methods.}
Deep learning methods, such as Graves' work, train RNN models conditioned on target characters \cite{graves2013generating,carter2016experiments,zhang2017drawing}. 
The intra-variance of a writer's style was achieved with Mixture Density Networks (MDN) as the final synthesis layer \cite{bishop1994mixture}. Berio et al.~use recurrent-MDN for graffiti style transfer \cite{berio2017calligraphic}.
However, these methods cannot learn to represent handwriting styles per writer, and so cannot perform writer style transfer. 

State-of-the-art models can generate characters in an inferred style \cite{Aksan:2018:DeepWriting}. Aksan et al.'s DeepWriting model uses Variational Recurrent Neural Networks (VRNN) \cite{VRNN_NIPS} and assumes a latent vector \emph{z} that controls writer handwriting style. 
Across writers, this method tends to average out specific styles and so reduces detail. 
Further, while sample efficient, VRNN models have trouble exploiting an abundance of inference samples because the style representation is only the last hidden state of an LSTM.
We avoid this limitation by extracting character-dependent style vectors from samples and querying them as needed in generation.

\paragraph{Sequence methods beyond handwriting.}
Learning-based architectures for sequences were popularized in machine translation \cite{cho-etal-2014-learning}, where the core idea is to encode sequential inputs into a fixed-length latent representation.
Likewise, text-to-speech processing has been improved by sequence models \cite{oord2016wavenet,shen2018natural}, with extensions to style representation for speech-related tasks like speaker verification and voice conversion. 
Again, one approach is to use the (converted) last output of an LSTM network as a style representation \cite{heigold2016end}. 

%
%
Other approaches \cite{hsu2018scalable,hsu2017learning} models multiple stylistic latent variables in a hierarchical manner and introduces an approach to transfer styles within a standard VAE setting \cite{kingma2013auto}. 

Broadly, variational RNN approaches \cite{Aksan:2018:DeepWriting,VRNN_NIPS,hsu2018scalable} have the drawback that they are incapability of improving generation performance with more inference samples. While VRNNs are sample efficient when only a few samples are available for style inference, a system should also generate better results as more inference samples are provided (as in \cite{mytext}). Our method attempts to be scalable and sample efficient through learning decoupled underlying generation factors.

We compare properties of four state of the art handwriting synthesis models (Tab.~\ref{table:comparison}), and illustrate two of their different approaches (Fig.~\ref{fig:his_she}).

\definecolor{good}{rgb}{0.75,0.9,0.75}
\definecolor{decent}{rgb}{0.9,0.93,0.75}
\definecolor{bad}{rgb}{0.9,0.75,0.75}
\definecolor{na}{rgb}{0.8,0.8,0.8}

\begin{table*}[t]
\centering
\caption{Property comparison of state-of-the-art handwriting generation models.}
\resizebox{\linewidth}{!}{
\begin{tabular}{l cccccccc}
\toprule
& 
Style& 
No human& 
Infinite& 
Synthesize mis-& 
Benefit from& 
Smooth&
Learn new
\\
Method & 
transfer? & 
segmentation? & 
variations? & 
sing samples? & 
more samples?& 
interpolation? &
characters?
\\
\midrule
Graves (2013) & \cellcolor{bad} No & \cellcolor{good} Yes & \cellcolor{good} Yes & \cellcolor{bad} No & \cellcolor{bad} No & \cellcolor{bad} No & \cellcolor{bad} No\\
Berio et al. (2017) & \cellcolor{good} Yes & \cellcolor{good} Yes & \cellcolor{good} Yes & \cellcolor{bad} No & \cellcolor{bad} No & \cellcolor{decent} Sort of & \cellcolor{bad} No\\
Haines et al. (2017) & \cellcolor{good} Yes & \cellcolor{bad} No & \cellcolor{decent} Sort of & \cellcolor{bad} No  & \cellcolor{good} Yes & \cellcolor{bad} No & \cellcolor{bad} No\\
Aksan et al. (2018) & \cellcolor{good} Yes & \cellcolor{decent} Sort of & \cellcolor{good} Yes & \cellcolor{good} Yes  & \cellcolor{bad} No & \cellcolor{decent} Sort of & \cellcolor{bad} No\\
Ours & \cellcolor{good} Yes & \cellcolor{good} Yes & \cellcolor{good} Yes & \cellcolor{good} Yes & \cellcolor{good} Yes & \cellcolor{good} Yes & \cellcolor{good} Yes\\
\bottomrule
\end{tabular}
}
\label{table:comparison}
\end{table*}

\section{Method}
\label{sec:method}

\paragraph{Input, preprocess, and output.}
A stroke sequence $\mathbf{x} = (p_1, \dots, p_N)$ has each $p_t$ store the change in $x$- and $y$-axis from the previous timestep ($\Delta x_t = x_t - x_{t-1}$, $\Delta y_t = y_t - y_{t-1}$), and a binary termination flag for the `end of stroke' ($eos = \{0, 1\}$). This creates an $(N,3)$ matrix. A character sequence $\mbs = (\mbc_1, \dots, \mbc_M)$ contains character vectors $\mbc_t$ where each is a one-hot vector of length equal to the total number of characters considered. This similarly is an $(M,Q)$ matrix.

The IAM dataset \cite{marti2002iam} and our stroke dataset were collected by asking participants to naturally write character sequences or words, which often produces cursive writing. As such, we must solve a segmentation problem to attribute stroke points to specific characters in $\mbs$. This is complex; we defer explanation to 
our supplemental.
For now, it is sufficient to say that we use unsupervised learning to train a segmentation network $k_{\theta}(\mathbf{x},\mathbf{s})$ to map regions in $\mathbf{x}$ to characters, and to demark `end of character' labels ($eoc = \{0, 1\}$) for each point.

As output, we wish to predict $\mathbf{x}'$ comprised of $\p'_t$ with: 1) coefficients for Mixture Density Networks \cite{bishop1994mixture} ($\pi_t, \mu_x, \mu_y, \sigma_x, \sigma_y, \rho$), which provide variation in output by sampling $\Delta x_t$ and $\Delta y_t$ from these distributions at runtime; 2) `end of stroke' $eos$ probability; and 3) `end of character' $eoc$ probability. This lets us generate cursive writing when $eos$ probability is low and $eoc$ probability is high.

\begin{figure}[t]
	\centering
	\includegraphics[width=\linewidth]{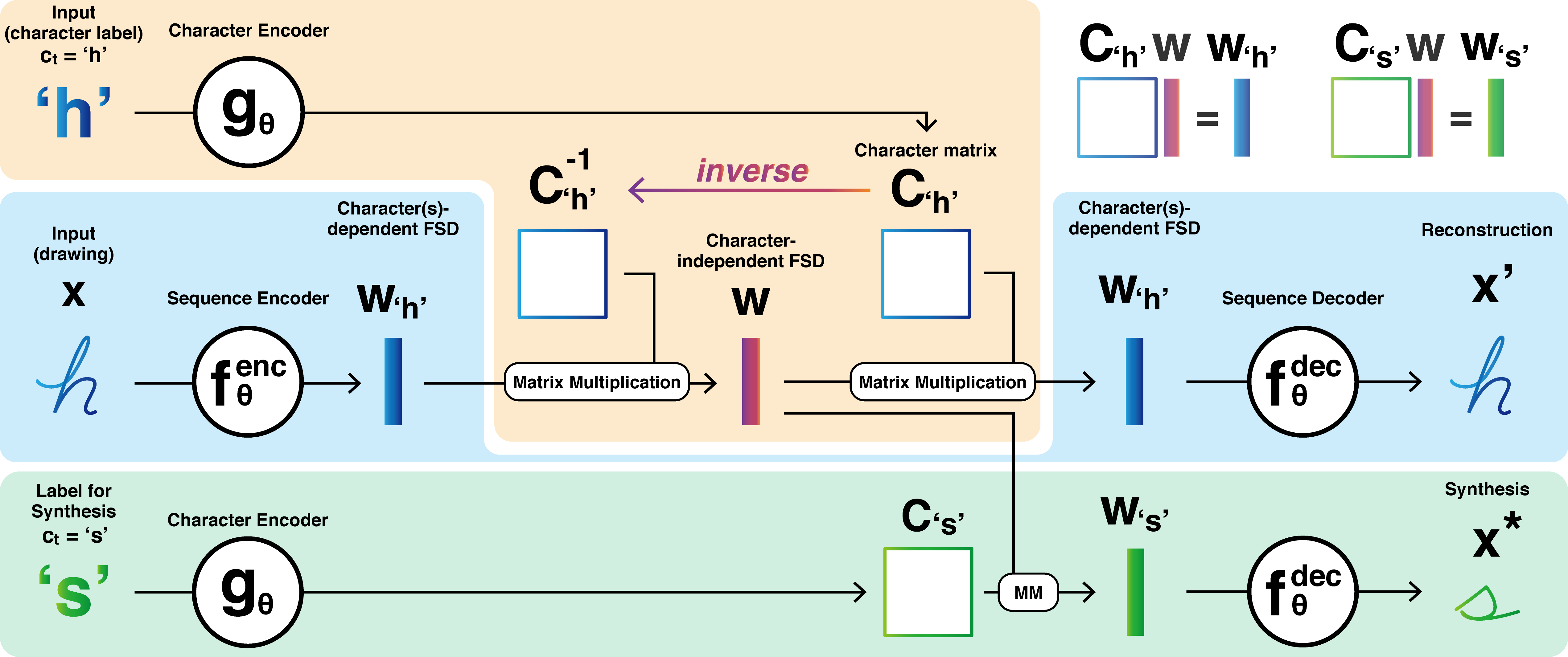}
    \caption{High-level architecture. Circles are parametrized function approximators, and rectangles/squares are variables. \emph{Blue region}: Encoder-decoder architecture. \emph{Orange region}: Character-conditioned layers. \emph{Green region}: Synthesis procedure.}
    \label{fig:general}
\end{figure}

\paragraph{Decoupled Style Descriptors (DSD).}
We begin with the encoder-decoder architecture proposed by Cho et al.~\cite{cho-etal-2014-learning}  (Fig.~\ref{fig:general}, blue region). Given a supervised database $\mathbf{x},\mathbf{s}$ and a target string $c_t$, to represent handwriting style we train a parameterized encoder function $f_{\theta}^{\text{enc}}$ to learn writer-dependent character-dependent latent vectors $\mbw_{c_t}$. Then, given $\mbw_{c_t}$, we simultaneously train a parameterized decoder function $f_{\theta}^{\text{dec}}$ to predict the next point $p'_t$ given all past points $p'_{1:t-1}$. Both encoder and decoder $f_{\theta}$ are RNNs such as LSTM models:
\begin{equation}
	p'_t = f_{\theta}^{\text{dec}}(p'_{1:t-1}|\mbw_{c_t}).
    \label{eq:1}
\end{equation}

This method does not factor character-independent writer style; yet, we have no way of explicitly describing this property via supervision and so we must devise a construction to learn it implicitly. Thus, we add a layer of abstraction (Fig.~\ref{fig:general}, orange region) with three assumptions: 
\begin{enumerate}
    \item If two stroke sequences $\mbx_1$ and $\mbx_2$ are written by the same writer, then consistency in their writing style is manifested by a character-independent writer-dependent latent vector $\mbw$. 
    \item If two character sequences $\mbs_1$ and $\mbs_2$ are written by different writers, then consistency in their stroke sequences is manifested by a character-dependent writer-independent latent matrix $\mbC$. $\mbC$ can be estimated via a parameterized encoder function $g_{\theta}$, which is also an RNN such as an LSTM:
        \begin{align}
        \mbC_{c_t} &= g_{\theta}(\mbs,c_t). \label{eq:genc}
        \end{align}
    \item $\mbC_{c_t}$ instantiates a writer's style $\mbw$ to draw a character via $\mbw_{c_t}$, such that $\mbC_{c_t}$ and $\mbw$ are latent factors:
        \begin{align}
            \mbw_{c_t} &= \mbC_{c_t}\mbw, \label{eq:chdepw} \\
            \mbw &= \mbC_{c_t}^{-1}\mbw_{c_t}. \label{eq:inv}
        \end{align}
\end{enumerate}

This method assumes that $\mbC_{c_t}$ is invertible, which we will demonstrate in Sec.~\ref{sec:experiments}. Intuitively, the multiplication of writer-dependent character vectors $\mbw_{c_t}$ with the inverse of character-DSD $\mbC_{c_t}^{-1}$ (Eq.~\ref{eq:inv}) factors out character-dependent information from writer-dependent information in $\mbw_{c_t}$ to extract a writer style representation $\mbw$. Likewise, Eq.~\ref{eq:chdepw} restores writer-dependent character $\mbw_{c_t}$ by multiplying the writer-specific style $\mbw$ with a relevant character-DSD $\mbC_{c_t}$. 

We use this property in synthesis (Fig.~\ref{fig:general}, green region). Given a target character $c_t$, we use encoder $g_\theta$ to generate a $\mbC$ matrix. Then, we multiply $\mbC_{c_t}$ by a desired writer style $\mbw$ to generate $\mbw_{c_t}$. Finally, we use trained decoder $f_{\theta}^{\text{dec}}$ to create a new point $p'_t$ given previous points $p'_{1:t-1}$:
\begin{align}
    p'_t &= f_{\theta}^{\text{dec}} (p'_{1:t-1} | \mbw_{c_t}), \text{    where } \mbw_{c_t} = \mbC_{c_t}\mbw.  \label{eq:synthesis}
\end{align}

\paragraph{Interpreting the linear factors.}
Eq.~\ref{eq:chdepw} states a linear relationship between $\mbC_{c_t}$ and $\mbw$. This exists at the latent representation level: $\mbw_{c_t}$ and $\mbC_{c_t}$ are separately approximated by independent neural networks $f_{\theta}^{\text{enc}}$ and $g_\theta$, which themselves are nonlinear function approximators~\cite{jia2016dynamic,wang2017zm}.
As $\mbC_{c_t}$ maps a vector $\mbw$ to another vector $\mbw_{c_t}$, we can consider $\mbC_{c_t}$ to be a fully-connected neural network layer (without bias). 
However, unlike standard layers, $\mbC_{c_t}$'s weights are not implicitly learned through backpropagation but are predicted by a neural network $g_{\theta}$ in Eq.~\ref{eq:genc}.
A further interpretation of $\mbC_{c_t}$ and $\mbC^{-1}_{c_t}$ as two layers of a network is that they respectively share a set of weights and their inverse.
Explicitly forming $\mbC_{c_t}$ in this linear way makes it simple to estimate $\mbC_{c_t}$ for \emph{new} characters that are not in the training dataset, given few sample pairs of $\mbw_{c_t}$ and $\mbw$, using standard linear least squares methods (Sec.~\ref{sec:newcharacters}).

\paragraph{Mapping character and stroke sequences with $f_\theta$ and $g_\theta$.}
Next, we turn our attention to how we map sequences of characters and strokes within our function approximators. Consider the LSTM $f_{\theta}^{\text{enc}}$: Given a character sequence $\mbs$ as size of $(M,Q)$ where $M$ is the number of characters, and a stroke sequence $\mbx$ of size $(N,3)$ where $N$ is the number of points, our goal is to obtain a style vector for each character $\mbw_{c_t}$ in that sequence. The output of our segmentation network $k_{\theta}$ preprocess defines `end of character' bits, and so we know at which point in $\mbx$ that a character switch occurs, e.g., from \textit{h} to \textit{e} in \textit{hello}. 

\begin{figure}[t]
	\centering
	\includegraphics[width=0.9\textwidth]{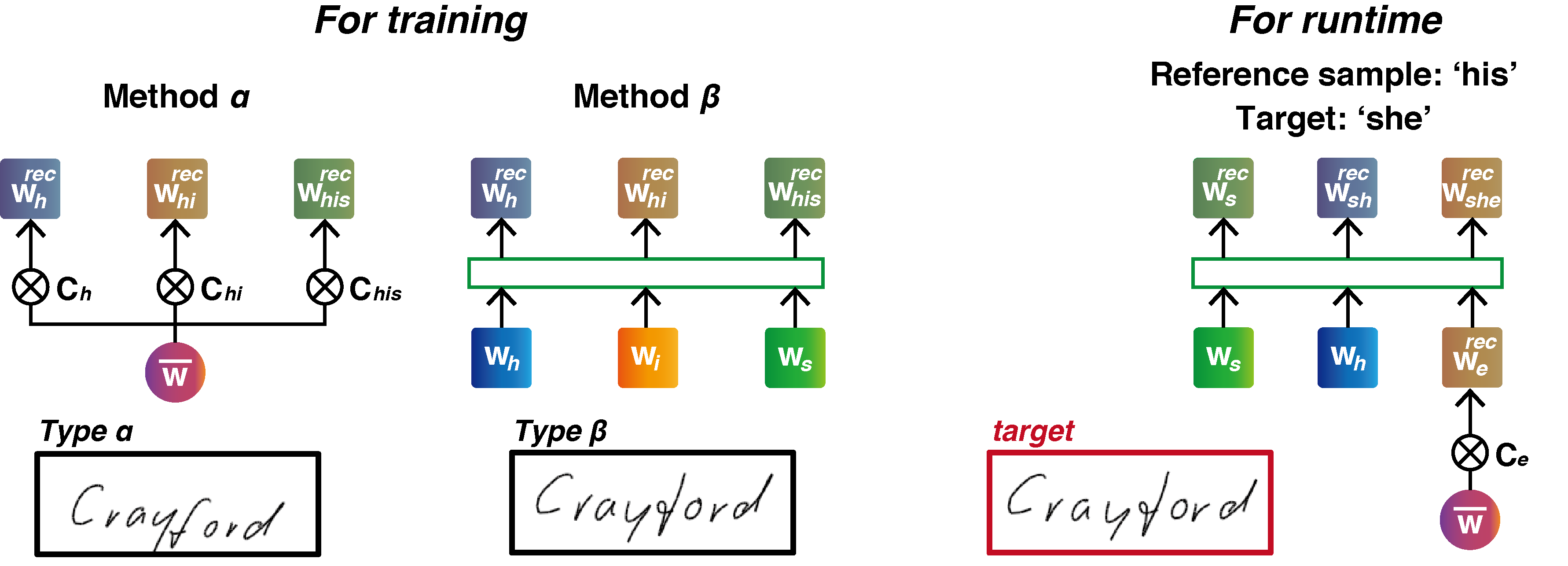}
	\caption{Reconstruction methods to produce writer-character-DSD $\mbw_{c_t}$, with training sample $\mbs,\mbx$ of \textit{his} and test sample $\mbs$ of \textit{she}. Green rectangle is $h_{\theta}$ as defined in Equation \ref{eq:9d}. \emph{Training}: Method $\alpha$ multiplies writer style $\overline{\mbw}$ with each character string matrix $\mbC_{c_t}$.
	Method $\beta$ restore temporal dependencies of segment-level writer-character-DSDs ($\mbw_h$, $\mbw_i$, $\mbw_s$) via an LSTM, which produces higher-quality results that are preferred by users (Sec.~\ref{sec:qualitativeeval}). Target test image is in red. \emph{Runtime}: Both prediction model Method $\alpha$ and $\beta$ are combined to synthesize a new sample given contents within the reference sample.
	}
    \label{fig:his}
\end{figure}

First, we encode $\mbx$ using $f_{\theta}^{\text{enc}}$ to obtain a $\mbx^*$ of size $(N, L)$, where $L$ is the latent feature dimension size (we use 256). Then, from $\mbx^*$, we extract $M$ vectors at these switch indices---these are our writer-dependent character-dependent DSDs $\mbw_{c_t}$. As $f_{\theta}^{\text{enc}}$ is an LSTM, the historical sequence data up to that index is encoded within the vector at that index (Fig.~\ref{fig:his}, top). For instance, for \textit{his}, $\mbx^*$ at switch index 2 represents how the writer writes the first two characters \textit{hi}, i.e., $\mbw_{hi}$. We refer to these $\mbw_{c_t}$ as `writer-character-DSDs'.

Likewise, LSTM $g_{\theta}$ takes a character sequence $\mbs$ of size $(M, Q)$ and outputs an array of $\mbC$ matrices that forms a tensor of size $(M, L, L)$ and preserves sequential dependencies between characters: The $i$-th element of the tensor $\mbC_{c_i}$ is a matrix of size $(L, L)$---that is, it includes information about previous characters up to and including the $i$-th character. 
Similar to $\mbx^*$, for \textit{his}, the second character matrix $\mbC_{c_2}$ contains information about the first two characters \textit{hi}---$\mbC$ is really a character sequence matrix.
Multiplying character information $\mbC_{c_t}$ with writer style vector $\mbw$ creates a writer-character-DSD $\mbw_{c_t}$.

\paragraph{Estimating $\mbw$.}
\label{sec:estimatingw}
\begin{wrapfigure}{r}{0.25\textwidth}
\vspace{-42pt}
\centering
\includegraphics[width=0.24\textwidth]{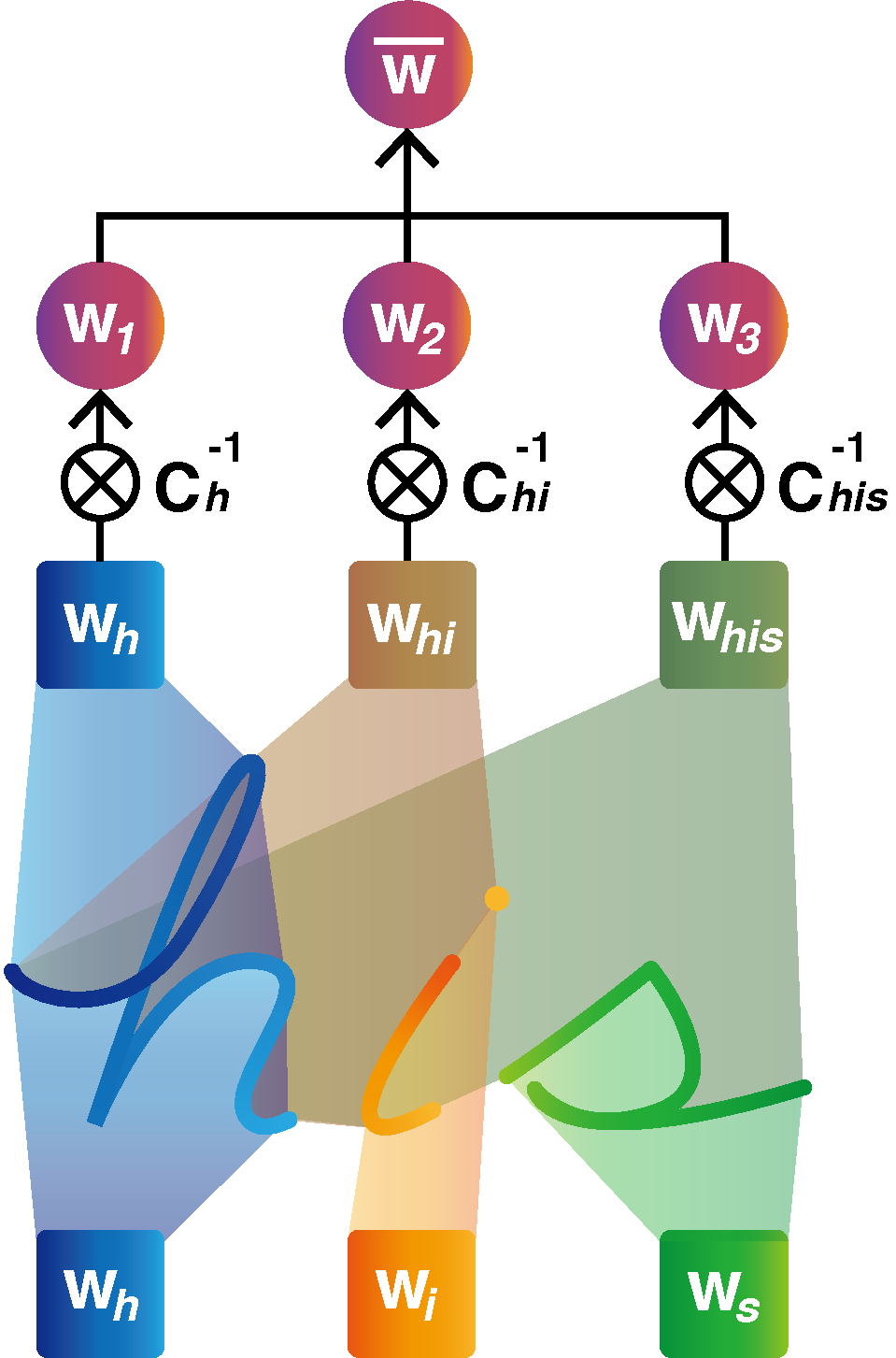}
\label{fig:mean_W}
\vspace{-30pt}
\end{wrapfigure}
When we encode a stroke sequence $\mbx$ that draws $\mbs$ characters via $f_{\theta}^{\text{enc}}$, we extract $M$ character(s)-dependent DSDs $\mbw_{c_t}$ (e.g., $\mbw_h$, $\mbw_{hi}$ and $\mbw_{his}$, \emph{right}). Via Eq.~\ref{eq:inv}, we obtain $M$ distinct candidates for writer-DSDs $\mbw$. To overcome this, for each sample, we simply take the mean to form $\mbw$:
\begin{equation}
\overline{\mbw} = \frac{1}{M} \sum_{t=1}^{M}{\mbC_{c_t}^{-1}\mbw_{c_t}}.
\label{eq:8}
\end{equation}

\paragraph{Generation approaches via $\mbw_{c_t}$.}
\label{sec:generationapproaches}
Consider the synthesis task in Fig.~\ref{fig:his_she}: given our trained model, generate how a new writer would write \textit{she} given a reference sample of them writing \textit{his}. 
From the \textit{his} sample, we can extract 1) segment-level writer-character-DSDs ($\mbw_h$, $\mbw_i$, $\mbw_s$), and 2) the global $\overline{\mbw}$.
To synthesize \textit{she}, our model must predict three writer-character-DSDs ($\mbw_s, \mbw_{sh}, \mbw_{she}$) as input to the decoder $f_{\theta}^{\text{dec}}$.
We introduce two methods to estimate $\mbw_{c_t}$:
\begin{subequations}
\begin{align}
\text{Method }\alpha: \mbw_{c_t}^{\alpha} &= \mbC_{c_t} \overline{\mbw} \label{eq:9a}\\
\text{Method }\beta: \mbw_{c_t}^{\beta} &= h_{\theta}([\mbw_{c_1}, \dots, \mbw_{c_{t}}])\label{eq:9d}
\end{align}
\end{subequations}
where $h_{\theta}$ is an LSTM that restore dependencies between temporally-separated writer-character-DSDs as illustrated in Fig.~\ref{fig:his}, green rectangle. We train our model to reconstruct $\mbw_{c_t}$ both ways. This allows us to use method $\alpha$ when test reference samples do not include target characters, e.g., \textit{his} is missing an \textit{e} for \textit{she}, and so we can reconstruct $\mbw_e$ via $\mbw$ and $\mbC_e$ (Fig.~\ref{fig:his}, right). It also allows us to use Method $\beta$ when test reference samples include relevant characters that, via $f_\theta^{\text{enc}}$, provide writer-character-DSDs, e.g., \textit{his} contains \textit{s} and \textit{h} in \textit{she} and so we can estimate $\mbw_s$ and $\mbw_h$. As these characters could come from any place in the reference samples, $h_{\theta}$ restores the missing sequence dependencies.

\subsection{Training losses}
\label{sec:loss}

We defer full architecture details for our supplemental material, and here explain our losses. We begin with a point location loss $\mathscr{L}^{loc}$ on predicted shifts in $x, y$ coordinates, $(\Delta x, \Delta y)$. As we employ mixture density networks as a final prediction layer in $f_{\theta}^{\text{dec}}$, we try to maximize the probability for the target shifts $(\Delta x^*, \Delta y^*)$ as explained by Graves et al.~\cite{graves2013generating}: 
$$\mathscr{L}^{loc}	= - \sum_{t} \log \Big(\sum_{j}\pi_t^j \mathcal{N}(\Delta x_t^*, \Delta y_t^* | {\mu_x}_t^j, {\mu_y}_t^j, {\sigma_x}_t^j, {\sigma_y}_t^j, \rho_t^j)\Big).$$
Further, we consider `end of sequence' flags $eos$ and `end of character' flags $eoc$ by computing binary cross-entropy losses $\mathscr{L}^{eos}, \mathscr{L}^{eoc}$ for each.

Next, we consider consistency in predicting writer-DSD $\mbw$ from different writer-character-DSDs $\mbw_{c_t}$. We penalize a loss $\mathscr{L}^{\mbw}$ that minimizes the variance in $\mbw_t$ in Equation \ref{eq:8}:
\begin{equation}
    \mathscr{L}^{\mbw} = \sum_t (\overline{\mbw} - \mbw_t)^2
    \label{eq:12}
\end{equation}

Further, we penalize the reconstruction of each writer-character-DSD. We compare the writer-character-DSD retrieved by $f_{\theta}^{\text{enc}}$ from inference samples as $\mbw_{c_t}$ to their reconstructions ($\mbw_{c_t}^{\alpha}$, $\mbw_{c_t}^{\beta}$) via generation Methods $\alpha$ and $\beta$:
\begin{equation}
	\mathscr{L}^{\mbw_{c_t}}_{A\in (\alpha, \beta)} = \sum_t (\mbw_{c_t} - \mbw_{c_t}^{A})^2
	\label{eq:11}
\end{equation}
When $t=1$, $\mathscr{L}_{\beta}^{\mbw_{c_1}} = (\mbw_{c_1} - h_{\theta}(\mbw_{c_1}))^2$. As such, minimizing this loss prevents $h_{\theta}$ in generation Method $\beta$ from diluting the style representation $\mbw_{c_1}$ generated by $f_{\theta}^{\text{enc}}$ because $h_{\theta}$ is induced to output $\mbw_{c_1}$.

Each loss can be computed for three types of writer-character-DSD $\mbw_{c_t}$: those predicted by $f_{\theta}^{\text{enc}}$, Method $\alpha$, and Method $\beta$. These losses can also be computed at character, word, and sentence levels, e.g., for words:
\begin{align}
        \mathscr{L}_{word} = \sum_{A\in (f_{\theta}^{\text{enc}}, \alpha, \beta)} \Big(\mathscr{L}^{loc}_{A} + \mathscr{L}^{eos}_{A} + \mathscr{L}^{eoc}_{A} + \mathscr{L}^{\mbw}_{A} + \mathscr{L}^{\mbw_{c_t}}_{A}\Big).
\label{eq:13}
\end{align}
Thus, the total loss is: $\mathscr{L}_{total} = \mathscr{L}_{char} + \mathscr{L}_{word} + \mathscr{L}_{sentence}$.

\noindent $\mathscr{L}_{f_{\theta}^{\text{enc}}}^{\mbw_{c_t}} = 0$ by construction from Equation \ref{eq:11}; we include it here for completeness.

Sentence-level losses help to make the model predict spacing between words. 
While our model could train just with character- and word-level losses, this would cause a problem if we ask the model to generate a sentence from a reference sample of a single word. Training with $\mathscr{L}_{sentence}$ lets our model predict how a writer would space words based on their writer-DSD $\mbw$.

\paragraph{Implicit $\mbC$ inverse constraint.} Finally, we discuss how $\mathscr{L}^{\mbw_{c_t}}$ at the character level implicitly constrains character-DSD $\mbC$ to be invertible. If we consider a single character sample, then mean $\overline{\mbw}$ in Equation \ref{eq:8} is equal to $\mbC_{c_1}^{-1}\mbw_{c_1}$. In this case, as $\mbw_{c_t}^{\alpha} = \mbC_{c_t} \overline{\mbw}$ (Eq.~\ref{eq:9a}), $\mathscr{L}_{\alpha}^{\mbw_{c_t}}$ becomes:
\begin{equation}
    \mathscr{L}_{\alpha}^{\mbw_{c_t}} = (\mbw_{c_1} - \mbC_{c_1}\mbC_{c_1}^{-1} \mbw_{c_1})^2
    \label{eq:15}
\end{equation}
This value becomes nonzero when $\mbC$ is singular ($\mbC\mbC^{-1}\neq \mbI$), and so our model avoids non-invertible $\mbC$s.

\paragraph{Training through inverses.} As we train our network end-to-end, our model must backpropagate through $\mbC_{c_t}$ and $\mbC^{-1}_{c_t}$. As derivative of matrix inverses can be obtained with $\frac{d\mbC^{-1}}{dx} = -\mbC^{-1}\frac{d\mbC}{dx}\mbC^{-1}$, our model can train.

\section{Experiments}
\label{sec:experiments}

\paragraph{Dataset.}
Our new dataset---BRUSH---provides characteristics that other online English handwriting datasets do not, including the typical online English handwriting dataset IAM~\cite{marti2002iam}. 
First, we explicitly display a baseline in every drawing box during data collection. This enables us to create handwriting samples whose initial action is the $x,y$ shift from the baseline to the starting point. This additional information might also help improve performance in recognition tasks.

Second, our $170$ individuals wrote $488$ words \emph{in common} across $192$ sentences. 
This helps to evaluate handwriting models and observe whether $\mbw$ and $\mbC$ are decoupled: given a sample that failed to generate, we can compare the generated results of the same word across writers. If writer A failed but B succeeded, then it is likely that the problem is not with $\mbC$ representations but with either $\mbw$ or $\mbw_{c_t}$. If both A and B failed to draw the word but succeeding in generating other words, it is likely that $\mbC$ or $\mbw_{c_t}$ representations are to blame. We provide further details about our dataset and collection process in our supplemental material.

Third, for DeepWriting~\cite{Aksan:2018:DeepWriting} comparisons, we use their training and test splits on IAM that mix writer identities---i.e., in training, we see some data from every writer. For all other experiments, we use our dataset, where we split between writers---our 20 test writers have never been seen writing \emph{anything} in training.

\paragraph{Invertibility of $\mbC$.}
To compute $\mbw$ in Equation \ref{eq:inv}, we must invert the character-DSD $\mbC$. Our network is designed to make $\mbC$ invertible as training proceeds by penalizing a reconstruction loss for $\mbw_{c_{t}}$ and $\mbC_{c_t}\mbC^{-1}_{c_t}\mbw_{c_{t}}$ (Sec.~\ref{sec:loss}). 
To test its success, we compute $\mbC$s from our model for all single characters ($86$ characters) and character pairs ($86^2 = 7,396$ cases), and found $\mbC$ to have full rank in each case. Next, we test all possible 3-character-strings ($86^3 = 636,056$ cases). Here, there were $37$ rare cases with non-invertible $\mbC$s, such as \emph{1Zb} and \emph{6ak}.
In these cases, we can still extract two candidate $\mbw$ from the first two characters (e.g., \emph{1} and \emph{1Z} in the \emph{1Zb} sample) to complete generation tasks.

\paragraph{Qualitative evaluation with users.}
\label{sec:qualitativeeval}
\begin{figure}[t]
\centering
\includegraphics[width=\textwidth]{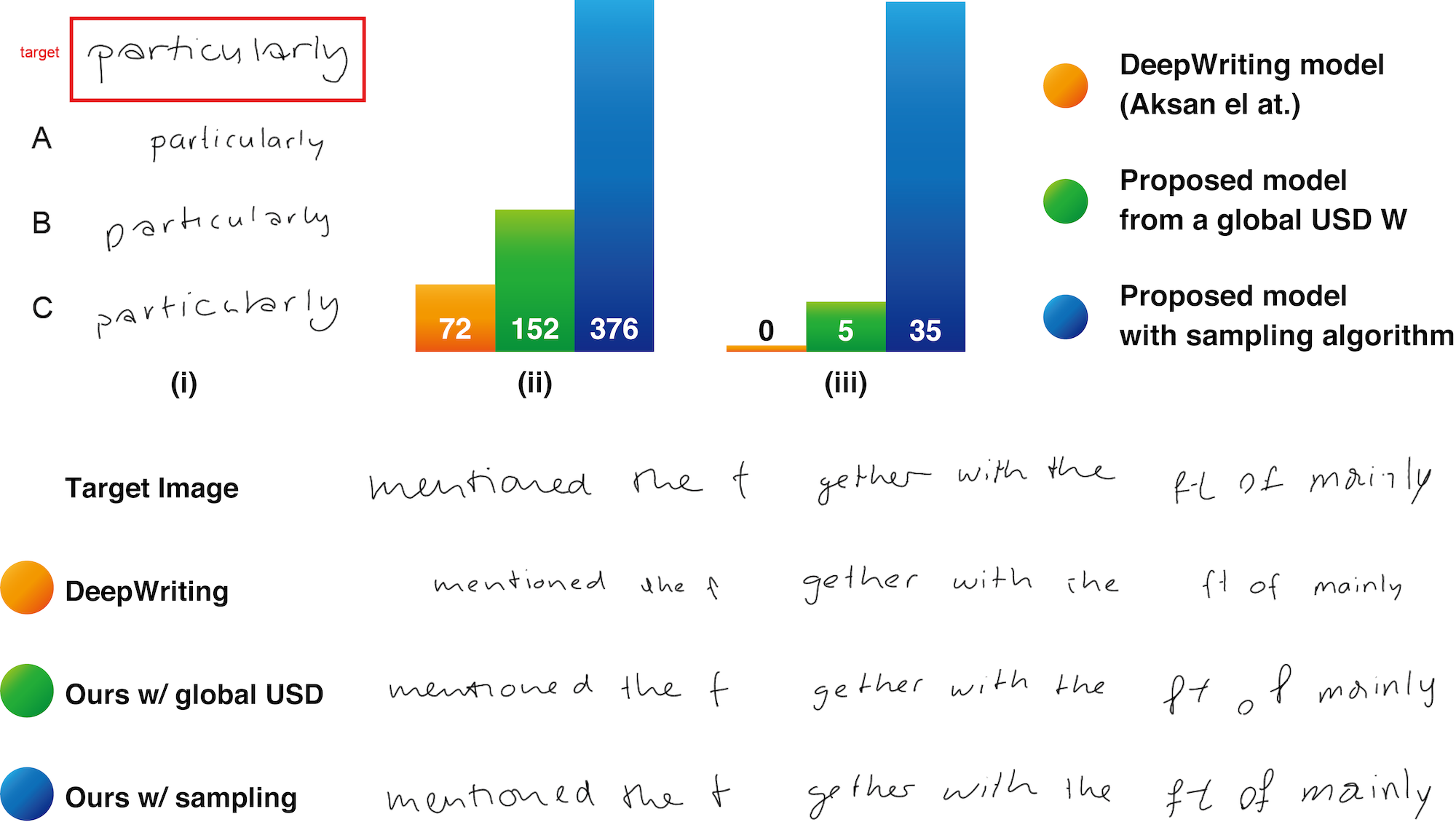}
\caption{Comparison of our proposed model vs.~the state-of-the-art model \cite{Aksan:2018:DeepWriting}. \emph{Top:} (i) Example writing similarity ordering task assigned to MTurk workers. (ii) Counts of most similar results with the target image. (iii) Sample-level vote. \emph{Bottom:} Three examples of task orderings; see supplemental for all 40. The model of Aksan et al.~\cite{Aksan:2018:DeepWriting} typically over-smooths the style and loses key details.}
\label{fig:score}
\end{figure}

We use Amazon Mechanical Turk to asked $25$ participants to rank generated handwriting similarity to a target handwriting (Fig.~\ref{fig:score} (i)).
We randomly selected $40$ sentence-level target handwriting samples from the validation set of IAM dataset \cite{marti2002iam}. Each participant saw randomly-shuffled samples; in total, $600$ assessments were made. 
We compared the abilities of three models to generate the same handwriting style without seeing the actual target sample.
We compare to the state-of-the-art DeepWriting model \cite{Aksan:2018:DeepWriting}, which uses a sample from the same writer (but of a different character sequence) for style inference. 
We test both Methods $\alpha$ and $\beta$ from our model. Method $\alpha$ uses the same sample to predict $\mbw$ and to generated a new sample. Method $\beta$ randomly samples $10$ sentence-level drawing by the target writer and creates a sample with the algorithm discussed in Sec.~\ref{sec:generationapproaches}. DeepWriting cannot take advantage of any additional character samples at inference time because it estimates only a single character-independent style vector.

Figure \ref{fig:score} (ii) displays how often each model was chosen as the most similar to the target handwriting; our model with sampling algorithm was selected $5.22\times$ as often as Aksan et al.'s model. Figure \ref{fig:score} (iii) displays which model was preferred across the $40$ cases: of the $15$ assessments per case, we count the number of times each model was the most popular. We show all cases in supplemental material.

\paragraph{Interpolation of $\mbw$, $\mbw_{c_t}$, and $\mbC$.}
\label{sec:interpolation}

\begin{figure}[t]
\centering
\includegraphics[width=\textwidth]{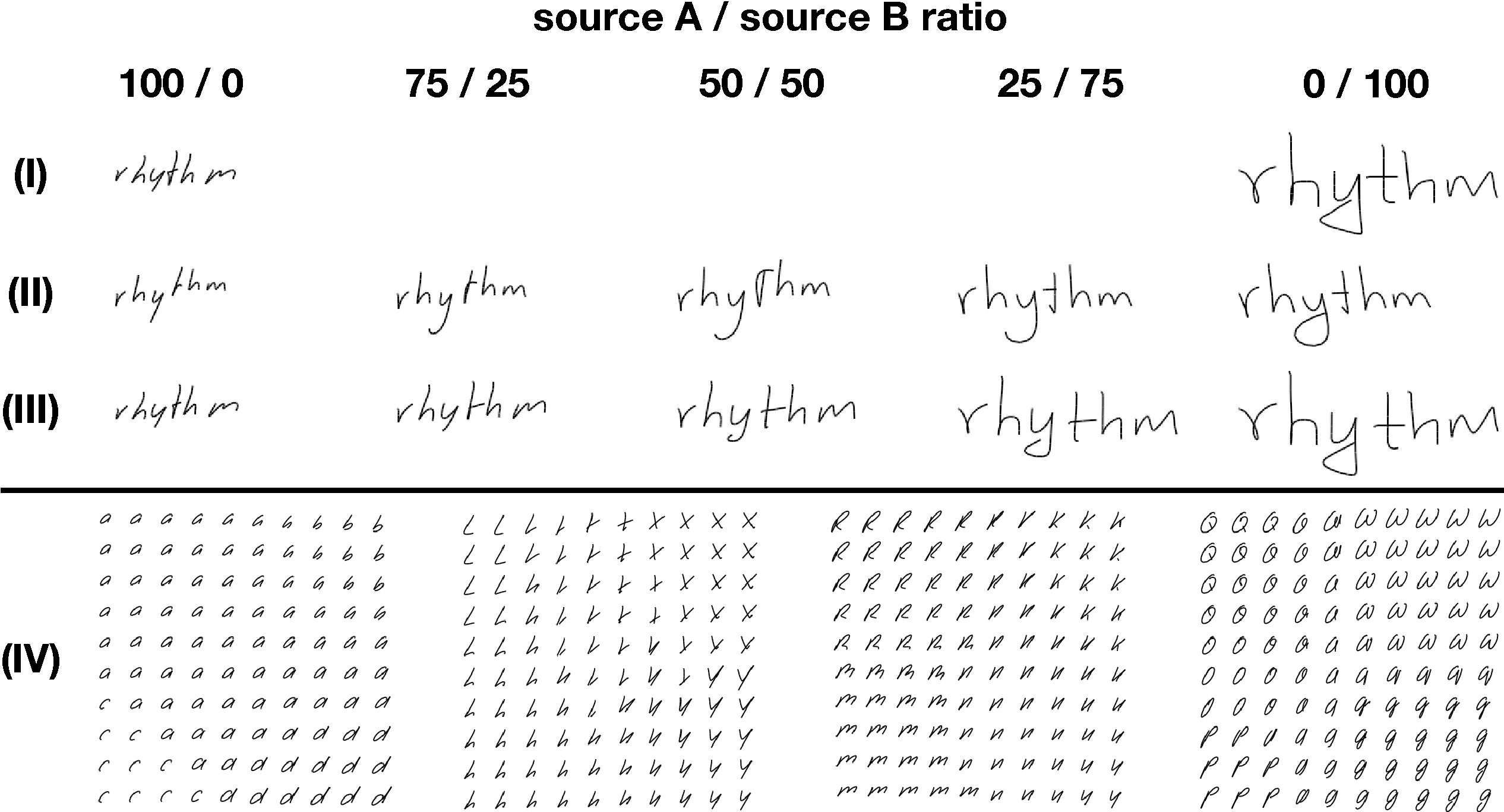}
\caption{Interpolation at different levels. (I) Original samples by two writers. (II) At the writer-DSD $\mbw$ level. (III) At the writer-character-DSD $\mbw_{c_t}$ level. (IV) At $\mbC$ level. \emph{Left to right:} Characters used are \emph{abcd}, \emph{Lxhy}, \emph{Rkmy}, \emph{QWPg}.}
\label{fig:interpolate}
\end{figure}

Figure \ref{fig:interpolate} demonstrates that our method can interpolate (II) at the writer-DSD $\mbw$ level, (III) at the writer-character-DSD $\mbw_{c_t}$ level, and (IV) at the character-DSD $\mbC$ level.
Given two samples of the same word by two writers $\mbx^A$ and $\mbx^B$, we first extract writer-character-DSDs from each sample (e.g., $\mbw_{rhy}^A$, $\mbw_{rhythm}^B$), then we derive writer-DSDs $\overline{\mbw^A}$ and $\overline{\mbw^B}$ as in Sec.~\ref{sec:estimatingw}. 
To interpolate by $\gamma$ between two writers, we compute the weighted average $\overline{\mbw^C}=\gamma\overline{\mbw^A} + (1-\gamma)\overline{\mbw^B}$. 
Finally, we reconstruct writer-character-DSDs from $\overline{\mbw^C}$ (e.g., $\mbw^{C}_{rhy} = \mbC_{rhy}\overline{\mbw^C}$) and feed this into $f^{dec}_{\theta}$ to generate a new sample. 
For (III), we simply interpolate at the sampled character-level (e.g., $\mbw_{rhy}^A$ and $\mbw_{rhy}^B$). 
For (IV), we bilinearly interpolate four character-DSDs $\mbC_{c_t}$ placed at the corners of each image: $\overline{\mbC} = (r_A \times \mbC_{A} + r_B \times \mbC_{B} + r_C \times \mbC_{C} + r_D \times \mbC_{D})$, where 
all $r$ sum to $1$.
From $\overline{\mbC}$, we compute a writer-character-DSD as  $\mbw_{\overline{c}} = \overline{\mbC}W$ and synthesize a new sample. In each case (II-IV), our representations are smooth.

\begin{figure}[t]
	\centering
	\includegraphics[width=\textwidth]{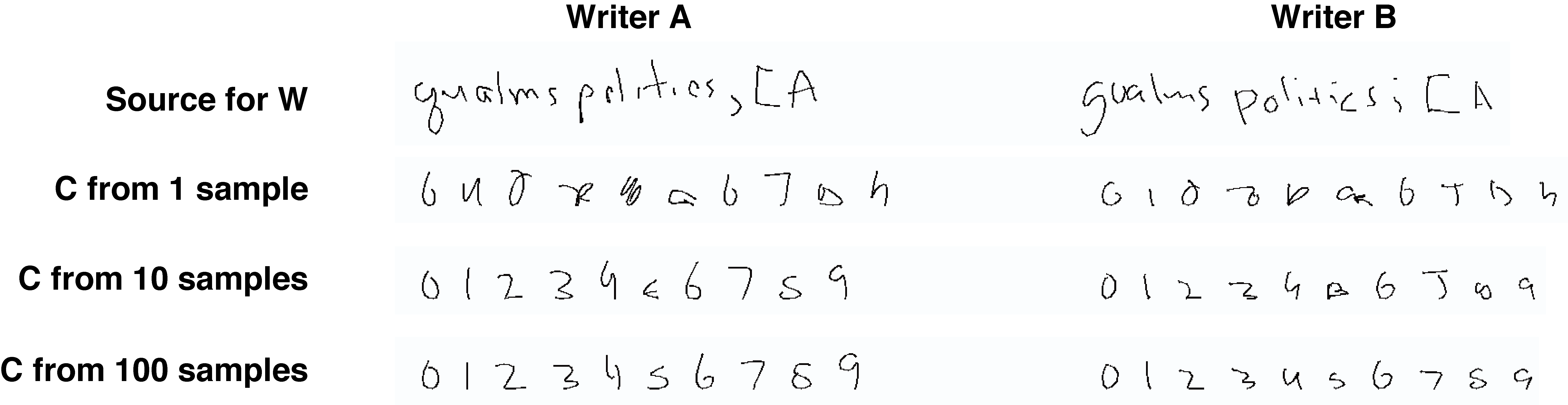}
    \caption{Predicting $\mbC$ from new character samples, given a version of our model that is not trained on numbers. As we increase the number of samples used to estimate $\mbC$, the better the stylistic differences are preserved when multiplying with $\mbw$s from different writers A and B. \emph{Note:} neither writers provided numeral samples; by our construction, samples can come from any writer.}
    \label{fig:new_chars}
\end{figure}

\paragraph{Synthesis of new characters.}
\label{sec:newcharacters}
Our approach allows us to generate handwriting for new characters from a few samples from any writer.  Let us assume that writer $A$ produces a new character sample \textit{3} that is not in our dataset. 
To make \textit{3} available for generation in other writer styles, we need to recover the character-DSD $\mbC_{\textit{3}}$ that represents the shape of the character \textit{3}. 
Given $\mbx$ for newly drawn character \textit{3}, encoder $f_\theta^{\text{enc}}$ first extracts the writer-character-DSD $\mbw_{\textit{3}}$. Assuming that writer $A$ provided other non-\textit{3} samples in our dataset, we can compute multiple writer-DSD $\mbw$ for $A$. This lets us solve for $\mbC_{\textit{3}}$ using least squares methods. We form matrices $\mbQ,\mbP_\textit{3}$ where each column of $\mbQ$ is one specific instance of $\mbw$, and where each column of $\mbP_\textit{3}$ is one specific instance of $\mbw_{\textit{3}}$. Then, we minimize the sum of the squared error, which is the Frobenius norm $\|\mbC_{\textit{3}}\mbQ-\mbP_{\textit{3}}\|^2_F$, e.g., via $\mbC_{\textit{3}}=\mbP_{\textit{3}}\mbQ^{+}$.

As architectured (and detailed in supplemental), $g_\theta$ actually has two parts: an LSTM encoder $g_{\theta}^{\text{LSTM}}$ that generates a 256$\times$1 character representation vector $\mbc_{c_t}^{\text{raw}}$ for a substring $c_t$, and a fully-connected layer $g_\theta^{\text{FC2}}$ that expands $\mbc^{\text{raw}}_{c_t}$ and reshapes it into a $256\times256$ matrix $\mbC_{c_t}=g_\theta^{\text{FC2}}(\mbc^{\text{raw}}_{c_t})$. Further, as the output of an LSTM, we know that $\mbc^{\text{raw}}_{c_t}$ should be constrained to values $[-1, +1]$. Thus, for this architecture, we directly optimize the (smaller set of) parameters of the latent vector $\mbc^{\text{raw}}_{c_t}$ to create $\mbC_{c_t}$ given the pre-trained fully-connected layer weights, using a constrained non-linear optimization algorithm (L-BFGS-B) via the objective %
$f(\mbc^{\text{raw}}_{c_t}) = \|\mbP_{\textit{3}} - g_\theta^{\text{FC2}}(\mbc^{\text{raw}}_{c_t})\mbQ\|^2_F$.

To examine this capability of our approach, we retrained our model with a modified dataset that \emph{excluded} numbers. In Figure \ref{fig:new_chars}, we see generation using our estimate of new $C$s from different sample counts. We can generate numerals in the style of a particular writer even though they never drew them, using relatively few drawing samples of the new characters from \emph{different} writers.

\paragraph{Writer recognition task.}
\label{sec:writerrecognition} 

Writer recognition systems try to assign test samples (e.g., a page of handwriting) to a particular writer given an existing database. Many methods use codebook approaches \cite{bensefia2002writer,bulacu2007text,he2015junction,bennour2019handwriting} to catalogue characteristic patterns such as graphemes, stroke junctions, and key-points from offline handwriting images and compare them to test samples. 
Zhang et al.~\cite{zhang2016end} extend this idea to online handwriting, and Adak et al.~study idiosyncratic character style per person and extract characteristic patches to identify the writer \cite{ch2019intravariable}. 

To examine how well our model might represent the latent distribution of handwriting styles, we perform a writer recognition task on our trained model on the randomly-selected $20$-writer hold out set from our dataset. 
First, we compute $20$ writer DSDs $\overline{\mbw}^{writer}_{i}$ from $10$ sentence-level samples---this is our offline `codebook' representing the style of each writer. Then, for testing, we sample from $1$--$50$ new word-level stroke sequences per writer (using words with at least 5 characters), and calculate the corresponding writer DSDs ($N=1,000$ in total). With the vector $L$ of true writer labels, we compute prediction accuracy:
\begin{equation}
    A = \frac{1}{N}\sum_{i=1}^{N}I(L_i, {\argmin_{j}(\overline{\mbw}_i^{word} - \overline{\mbw}_j^{writer})^2})
    \text{, } I(x,y) = \begin{cases}
        1, & \text{if } x = y\\
        0, & \text{otherwise}
    \end{cases}
    \label{eq:19}
\end{equation}
\begin{wrapfigure}{r}{0.41\textwidth}
\vspace{-22pt}
\centering
\includegraphics[width=0.4\textwidth]{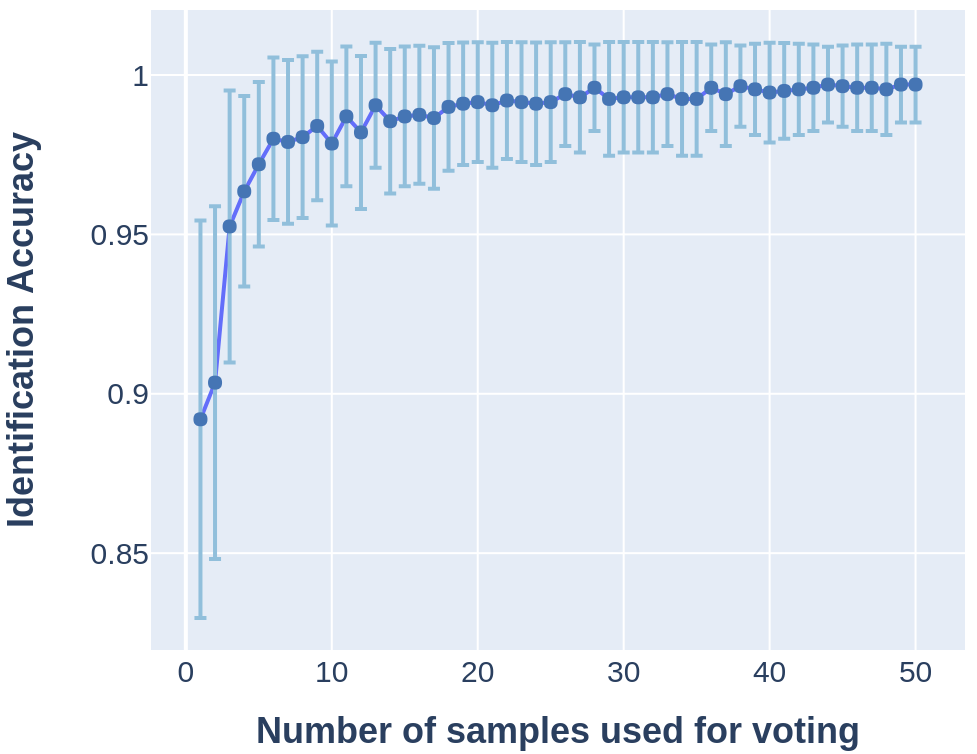}
\label{fig:accuracy}
\vspace{-20pt}
\end{wrapfigure}
We repeated the random sampling of $1$--$50$ words over $100$ trials and compute mean accuracy and standard error. When multiple test samples are provided, we predict writer identity for each word and average their predictions. Random accuracy performance is $5\%$. Our test prediction accuracy rises from {$89.20\% \pm 6.23$} for one word sample, to {$97.85\% \pm 2.57$} for ten word samples, to {$99.70\% \pm 1.18$} for 50 word samples. Increasing the number of test samples per writer increases accuracy because some words may not be as characteristic as others (e.g., `tick' vs. `anon'). Overall, while our model was not trained to perform this classification task, we can still achieved promising accuracy results from few samples---this is an indication that our latent space is usefully descriptive.

\paragraph{Additional experiments.} 
In our supplemental material, along with more architecture, model training procedure, and sampling algorithm details, we also: 1) compare to two style extraction pipelines, a stacked FC+ReLU layers and AdaIN, and find our approach more capable; 2) demonstrate the importance of learning style and content of character-DSD $\mathbf{C}$ by comparing with a randomly-initialized version; 3) ablate parts of our loss function, and illustrate key components; 4) experimentally show that our model is more efficient than DeepWriting by comparing generation given the same number of model parameters.

\section{Discussion}

While users preferred our model in our study, it still sometimes fails to generate readable letters or join cursive letters. One issue here is the underlying inconsistency in human writers, which we only partially capture in our data and represent in our model (e.g., cursive inconsistency). Another issue is collecting high-quality data with digital pens in a crowdsourced setting, which can still be a challenge and requires careful cleaning (see supplemental for more details).

\paragraph{Decoupling additional styles.}
Our model could potentially scale to more styles. For instance, we might create an age matrix $\mbA$ from a numerical age value $a$ as $\mbC$ is constructed from $c_t$, and extract character-independent age-independent style descriptor as $\mbw^* = \mbA^{-1} \mbC^{-1}_{c_t} \mbw_{c_t}$.
Introducing a new age operator $\mbA$ invites our model to find latent-style similarities across different age categories (e.g., between a child and a mature writer). Changing the age value and thus $\mbA$ may predict how a child's handwriting changes as s/he becomes older. However, training multiple additional factors in this way is likely to be challenging.

\paragraph{Alternatives to linear $\mbC$ multiplication operator.}
Our model can generate new characters by approximating a new $\mbC$ matrix from few pairs of $\mbw$ and $\mbw_{c_t}$ thanks to their linear relationship. However, one might consider replacing our matrix multiplication `operator' on $\mbC$ with parametrized nonlinear function approximators, such as autoencoders. Multiplication by $\mbC^{-1}$ would become an encoder, with multiplication by $\mbC$ being a decoder; in this way, $g_\theta$ would be tasked with predicting encoder weights given some predefined architecture. Here, consistency with $\mbw$ must still be retained. We leave this for future work.

\section{Conclusion}

We introduce an approach to online handwriting stroke representation via the Decoupled Style Descriptor (DSD) model. 
DSD succeeds in generating drawing samples which are preferred more often in a user study than the state-of-the-art model. 
Further, we demonstrate the capabilities of our model in interpolating samples at different representation levels, recovering representations for new characters, and achieving a high writer-identification accuracy, despite not being trained explicitly to perform these tasks. 
Online handwriting synthesis is still challenging, particularly when we infer a stylistic representation from few numbers of samples and try to generate new samples. 
However, we show that decoupling style factors has potential, and believe it could also apply to style-related tasks like transfer and interpolation in other sequential data domains, such as in speech synthesis, dance movement prediction, and musical understanding.

\paragraph{Acknowledgements.} This work was supported by the Sloan Foundation and the National Science Foundation under award number IIS-1652561. We thank Kwang In Kim for fruitful discussions and for being our matrix authority. We thank Naveen Srinivasan and Purvi Goel for the ECCV deadline snack delivery service. Finally, we thank all anonymous writers who contributed to our dataset.

\clearpage
\bibliographystyle{splncs04}
\bibliography{genwriting}
\vfill

\pagebreak
\appendix
\begin{minipage}[t][0.75cm][b]{0,5\textwidth}
\end{minipage}
\section*{\centering{Appendices for}\\\Large{Generating Handwriting via \\Decoupled Style Descriptors}}
\addcontentsline{toc}{section}{Appendices}
\renewcommand{\thesubsection}{\Alph{subsection}}


\vspace{0.5cm}
\contentsline {section}{\numberline {A}Table of Variables}{18}{section.A.1}%
\contentsline {section}{\numberline {B}Comparison with Style Transfer Baselines}{19}{section.A.2}%
\contentsline {section}{\numberline {C}Investigating the $\mathbf {C}$-matrix}{19}{section.A.3}%
\contentsline {section}{\numberline {D}Network Capacity}{23}{section.A.4}%
\contentsline {section}{\numberline {E}Further Generated Comparisons}{24}{section.A.5}%
\contentsline {section}{\numberline {F}Sampling Algorithm for Writer-Character-DSD $\mathbf {w}_{c_t}$}{27}{section.A.6}%
\contentsline {section}{\numberline {G}Sequence Decoder $f_{\theta }^{\text {dec}}$}{28}{section.A.7}%
\contentsline {section}{\numberline {H}Character Encoder Function $g_\phi $}{29}{section.A.8}%
\contentsline {section}{\numberline {I}Segmentation Network $k_{\theta }$}{30}{section.A.9}%
\contentsline {section}{\numberline {J}Detailed Training Procedure}{31}{section.A.10}%
\contentsline {section}{\numberline {K}Dataset Specification and Collection Methodology}{32}{section.A.11}%


\section{Table of Variables}

We include a brief table of the key variables used throughout the main manuscript and in this supplemental manuscript (Table~\ref{table:variables}).

\begin{table*}
\centering
\caption{Brief explanation of key variables used throughout these manuscripts.}
\resizebox{\linewidth}{!}{
\begin{tabular}{l l r l}
\toprule
&                   Name                    & Shape             & Note \\
\midrule
$\mathbf{x}$        & Input data            & (N, 3)            & A handwriting sample; a time sequence of 2D points.\\
$\mathbf{x}^*$        & Encoded input            & (N, 256)            & A raw output from $f_\theta^{\text{enc}}(\mbx)$. \\
$s$        & Sentence              & (M)               & A string label for $\mathbf{x}$ (e.g., \emph{hello}).\\
${c_t}$             & Substring            & (t)               & A substring of $s$ (e.g., \emph{he}).\\
$\mathbf{c}_t$			& Character vector	& (87$\times$1)		& A one-hot vector denoting the $t$-th character in $s$ .\\
$\mathbf{c}_t^{\text{raw}}$			& Encoded character	& (256$\times$1)		& An output from $g_\theta^{\text{FC1}}(\mathbf{c}_t)$. Input for $g_\theta^{\text{LSTM}}$.\\
$\mathbf{c}^{\text{raw}}_{c_t}$ &  Encoded substring	& (256$\times$1)		& An output from $g_\theta^{\text{LSTM}}(\mathbf{c}_{t}^{\text{raw}})$.\\
\midrule
$\mathbf{w}$        & Writer-DSD            & (256$\times$1)    & Content-independent handwriting style for a writer $A$.\\
$\mathbf{C}_{c_t}$  & Character-DSD         & (256$\times$256)  & An encoded character matrix for a substring $c_t$.\\
$\mathbf{w}_{c_t}$  & Writer-Character-DSD  & (256$\times$1)    & An encoded drawing representation for $c_t$, extracted from $\mbx^*$. \\
\midrule
$f_{\theta}^{\text{enc}}$ & Sequence encoder		&  & Outputs a list of Writer-Character-DSDs $\mathbf{w}_{c_t}$ from an input drawing $\mathbf{x}$.\\
$f_{\theta}^{\text{dec}}$ & Sequence decoder		&  & Outputs a drawing $\mathbf{x}$ from a list of $\mathbf{w}_{c_t}$.\\
$g_{\theta}$ & Character encoder		&  & Outputs a character matrix $\mathbf{C}_{c_t}$. Simplified function used as shorthand.\\
$g_{\theta}^{\text{FC1}}$ & 	&  & Outputs a vector $\mathbf{c}_t^{\text{raw}}$ from a vector $\mathbf{c}_t$.\\
$g_{\theta}^{\text{LSTM}}$ & &  & Outputs a vector $\mathbf{c}_{c_t}^{\text{raw}}$ from a list [$\mathbf{c}_1^{\text{raw}}$, ..., $\mathbf{c}_t^{\text{raw}}$].\\
$g_{\theta}^{\text{FC2}}$ & & & Outputs a Character-DSD $\mathbf{C}_{c_t}$ from a vector $\mathbf{c}_{c_t}^{\text{raw}}$.\\
$h_{\theta}$ & Temporal encoder && LSTM to restore dependencies between Writer-Character DSDs $\mathbf{w}_{c_t}$.\\
$k_{\theta}$ & Segmentation function && Segments a handwriting sample $\mathbf{x}$ into characters. \\
\bottomrule
\end{tabular}
}
\label{table:variables}
\end{table*}
\pagebreak

\section{Comparison with Style Transfer Baselines}
\label{sec:styletransferbaselines}

We evaluated our proposed model against two style transfer baselines. We define a style vector as $\mathbf{s} = f_{\theta}^{\text{enc}}(\mathbf{x})$ and a character-content vector as  $\mathbf{c} = g_{\theta}(c_t)$. 
To interweave $\mathbf{s}$ and $\mathbf{c}$, we consider a new operator $F$ where $F(\mathbf{s}, \mathbf{c})=z$. Then, we feed $z$ into our decoder function $f_{\theta}^{\text{dec}}$ to synthesize a drawing. We examined two operators for $F$: A) three stacked FC+ReLU layers, and B) AdaIN layer~\cite{Huang_2017_ICCV}.

In our method, $f_{\theta}^{\text{enc}}(\mathbf{x})$ produces $\mathbf{w}_{c_t}$, which is then decoupled from the character content via our $\mathbf{C}$ matrix operation. In Method A and B, $f_{\theta}^{\text{enc}}(\mathbf{x})$ produces $\mathbf{s}$, and via $F$ the network must implicitly represent content and writer style parts. For fairness, we keep the architectures of $f_{\theta}^{\text{enc}}$, $f_{\theta}^{\text{dec}}$, $g_{\theta}$ the same as in our approach, and train each method from scratch with the same data and loss function as in our approach.

\begin{figure}[t]
\centering
\includegraphics[width=1.0\linewidth]{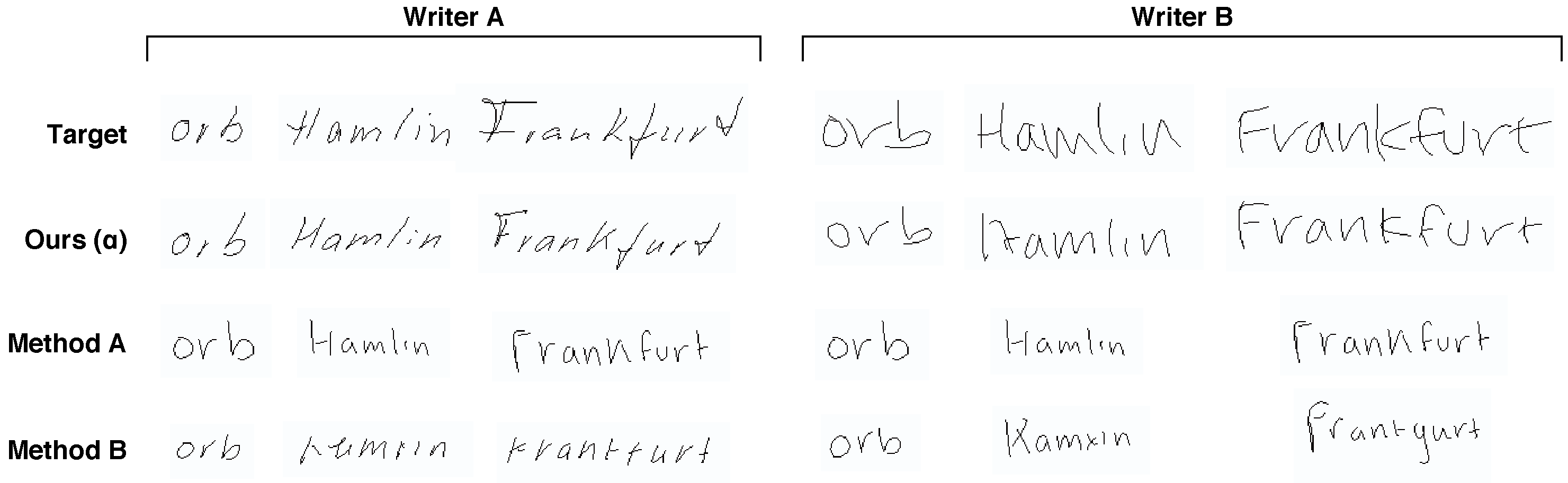}
\caption{Qualitative evaluation of two common style-transfer techniques.
}
\label{fig:style}
\end{figure}

Neither Method A or B is competitive with our method or with DeepWriting~\cite{Aksan:2018:DeepWriting}. While A and B can generate readable letters, A) has only one style, and B) fails to capture important character shape details leaving some illegible, and has only basic style variation like slant and size (Figure~\ref{fig:style}). This is because $f_{\theta}^{\text{enc}}$ must represent a content-independent style for a reference sample without its content information. The DeepWriting model decouples style and content by making $f_{\theta}^{\text{enc}}$ additionally predict the reference content via a character classification loss. Our approach does not try to decouple style and content within $f_{\theta}^{\text{enc}}$; instead our model extracts style from the output of $f_{\theta}^{\text{enc}}$ by multiplication with a content-conditioned matrix $\mathbf{C}$.

This simple experiment demonstrates that style-content decoupling is a difficult task. Instead of making one network (i.e., $f_{\theta}^{\text{enc}}$) responsible for filtering out content information from the style reference sample implicitly, we show empirically that our method to structurally decouple content information via $\mathbf{C}$ matrix multiplication is more effective in the online handwriting domain.

\section{Investigating the $\mathbf{C}$-matrix}

The $\mathbf{C}$ matrix for a character string $c_t$---$\textbf{C}_{c_t}$---is designed to contain information about how people generally write $c_t$: its role is to extract character(s)-specific information from $\textbf{w}_{c_t}$. Intuitively, the relationship between $\textbf{C}_{c_t}$ and $\textbf{w}_{c_t}$ can be seen as one of a key and a key-hole. Our model tries to create a perfect fit between a key ($\textbf{w}_{c_t}$) and a key-hole ($\textbf{C}_{c_t}$), where both shapes are learned simultaneously. But what if we fix the key-hole shape ahead of time, and simply learn to fit the key? That is, what if we assign pre-defined values to substring character matrices $\textbf{C}$ ahead of time? This would reduce the number of model parameters, speed up training and inference, and allow us to store $\mathbf{C}$ in memory as a look-up table rather than predict its values.

One issue with fixing the $\textbf{C}$ matrix is the exponential growth in the number of possible strings $c_t$ as we allow longer words. Thus, for this analysis, we will initialize $\textbf{C}$ for single- and two-character substrings only, which have a tractable number of variations in our Latin alphabet (Sec.~\ref{sec:dataset}). For example, instead of $\textbf{C}_{hello}$ for a word `hello', we consider its five constituent single- and two-character substrings $\textbf{C}_{h}$, $\textbf{C}_{he}$, $\textbf{C}_{el}$, $\textbf{C}_{ll}$, $\textbf{C}_{lo}$. Consequently, we modified the training data format by segmenting every sentence into two-character pairs.

%
%
\begin{figure}[p!]
\centering
\begin{subfigure}{.49\textwidth}
  \centering
  \includegraphics[width=\linewidth]{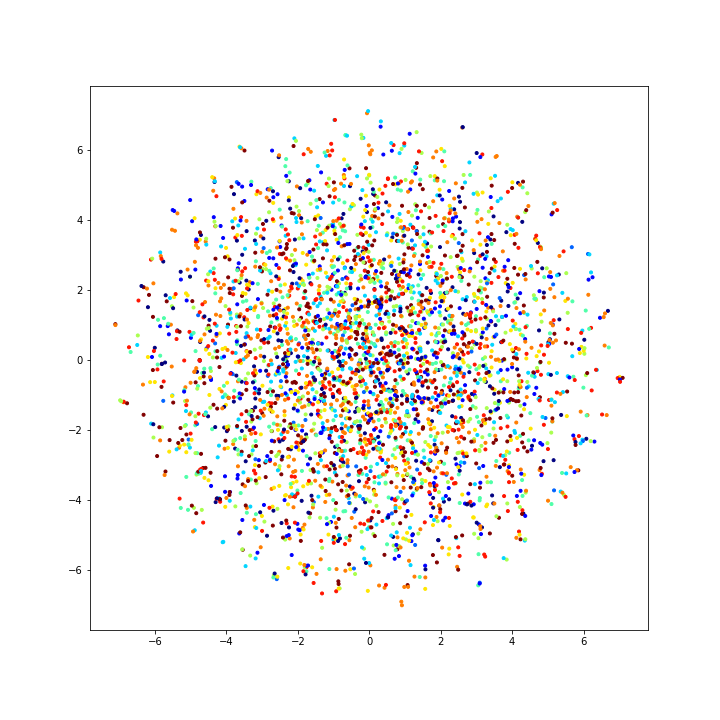}
  \caption{Random $\textbf{C}$ (2-character-string)}
  \label{fig:2a}
\end{subfigure}%
\begin{subfigure}{.49\textwidth}
  \centering
  \includegraphics[width=\linewidth]{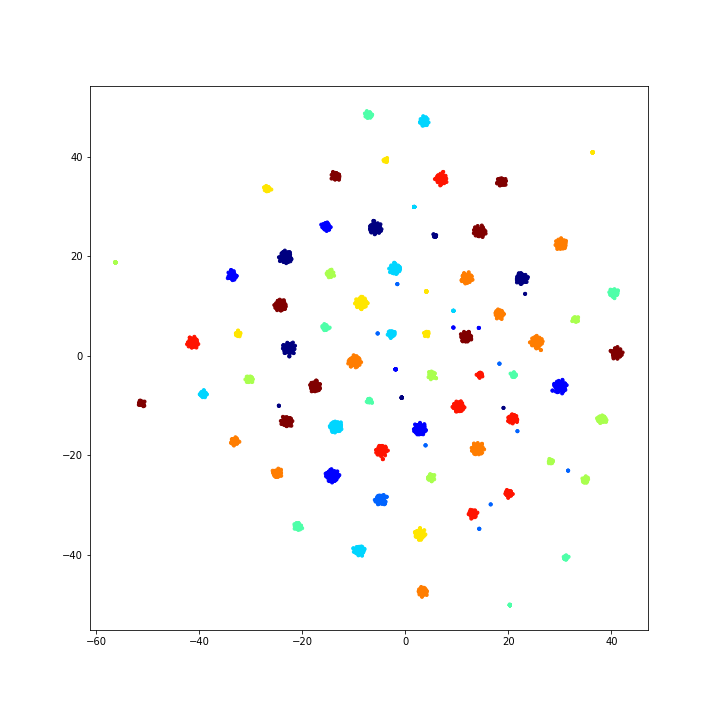}
  \caption{Spaced $\textbf{C}$ (2-character-string)}
  \label{fig:2b}
\end{subfigure}%
\vskip\baselineskip
\begin{subfigure}{.49\textwidth}
  \centering
  \includegraphics[width=\linewidth]{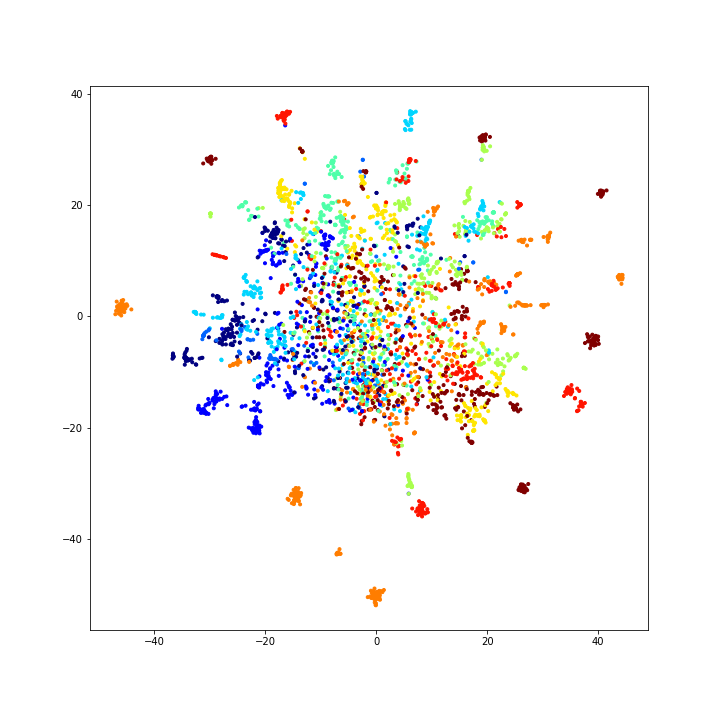}
  \caption{Learned $\textbf{C}$ (2-character-string)}
  \label{fig:2c}
\end{subfigure}
\begin{subfigure}{.49\textwidth}
  \centering
  \includegraphics[width=\linewidth]{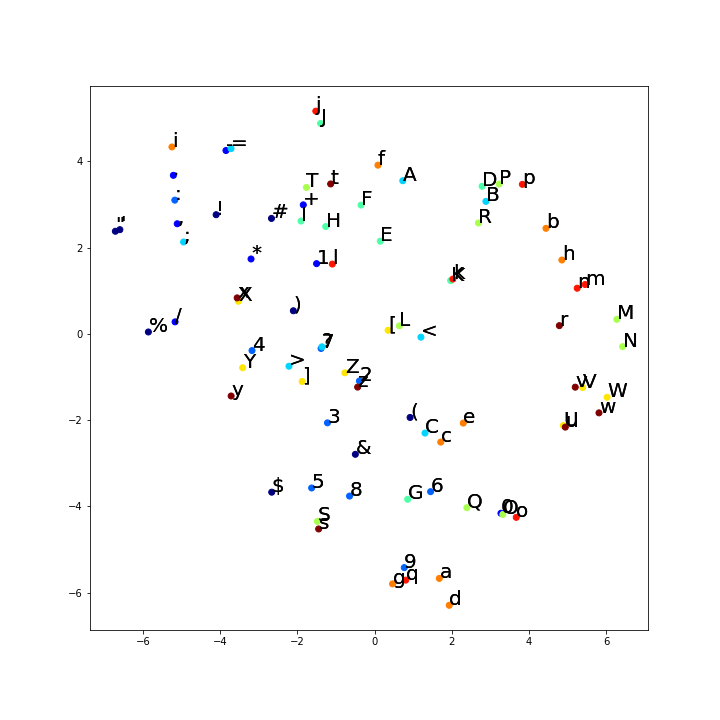}
  \caption{Learned $\textbf{C}$ (1-character-string)}
  \label{fig:2d}
\end{subfigure}
\caption{t-SNE visualization of different $\textbf{C}$. Each dot indicates different substring. The substrings with the same last character (e.g., `a\textbf{b}', `b\textbf{b}', `c\textbf{b}') are colored same. The learned $\textbf{C}$ in Figure \ref{fig:2c} are mostly concentrated in the middle. As each $\textbf{C}$ contains information about how to draw \emph{two} characters, even the two $\textbf{C}$ with the same last character (e.g., `a\textbf{b}' and `b\textbf{b}') are often distant from each other, because of the different first character. By contrast, isolating the single characters within the learned $\textbf{C}$ (Figure \ref{fig:2d}) shows them to be well mapped in the space: similar characters such as `\textbf{(}', `\textbf{c}' and `\textbf{C}' are closely positioned.}
\label{fig:2}
\end{figure}

We consider three scenarios (Figure \ref{fig:2}):
\begin{description}
    \item[Fixed random single- and two-character $\mbC$] In principle, each substring that $\textbf{C}$ represents only needs to be \emph{different} from other substrings, and so we assign a random matrix to each two-character substring.
    \item[Fixed well-spaced single- and two-character $\mbC$] Two matrices $\textbf{C}_{sh}$ and $\textbf{C}_{he}$ could contain mutual information about how to write the character $h$, and so we try to assign fixed matrices in a way that places similar substrings close to each other in high-dimensional space.
    \item[Learned single- and two-character $\mbC$] Our model trained only on single characters and two-character pairs. This trains $g_\theta$ to predict the values of $\mbC$.
\end{description}

\paragraph{Well-spaced C.} If we randomly initialize $\textbf{C}$ (i.e., $I(\textbf{C}_{sh}; \textbf{C}_{he}) \approx 0$), the values of two writer-character-DSDs, $\textbf{w}_{he}$ and $\textbf{w}_{she}$, must be significantly different from each other to output consistent $\textbf{w}$, and this makes the learning task harder for the $f_{\theta}^{\text{enc}}$ LSTM. Instead, to determine how to manually initialize $\textbf{C}$ such that they are more well spaced out, we look at the character information within $\textbf{w}_{c_t}$. As we use an LSTM to encode the input drawing to obtain $\textbf{w}_{c_t}$, it models long temporal dependencies. In other words, by the nature of LSTMs, $\textbf{w}_{c_t}$ tends to `remember' more recent characters than older characters, and so we assume $\textbf{w}_{c_t}$ remembers the second character more than the first character. Thus, we initialize the character matrix for a two-character substring $c_t$ as follows:
\begin{equation}
    \textbf{C}_{c_t} = r\textbf{C}_{c_1} + (1.0-r)\textbf{C}_{c_2}
\end{equation}
where $\textbf{C}_{c_1}$ and $\textbf{C}_{c_2}$ are randomly initialized single-character-DSDs, and we set $r = 0.1$. This leads to $\textbf{C}_{c_t}$ that have the same ending character (i.e., $c_2$) having similar representations, as shown in Figure \ref{fig:2b}.

\paragraph{Results.} Under t-SNE projections, the learned $\mbC$ models create more meaningful $\mbC$ representation layouts (Figure \ref{fig:2}). Unlike the well-spaced $\mbC$, when we project $\mbC$ for two-character substrings (Figure \ref{fig:2c}), we see a few outer clusters, with a larger `more chaotic' central concentration. As each dot in the projections represents $\mbC$ for two characters, these $\mbC$ cannot be easily clustered by the ending characters (e.g., considering general shapes, `c\textbf{b}' is likely to have a representation closer to the one of `C\textbf{6}' than `f\textbf{b}', despite the common 2nd character \textbf{b}). When looking at just the single-characters within our learned $\mbC$ representations (Figure \ref{fig:2d}), characters with similar shapes (e.g., `9',`q',`g') are closely positioned, and this indicates a successful representation learning for $\mbC$.

\begin{figure}[t]
\includegraphics[width=1.0\linewidth]{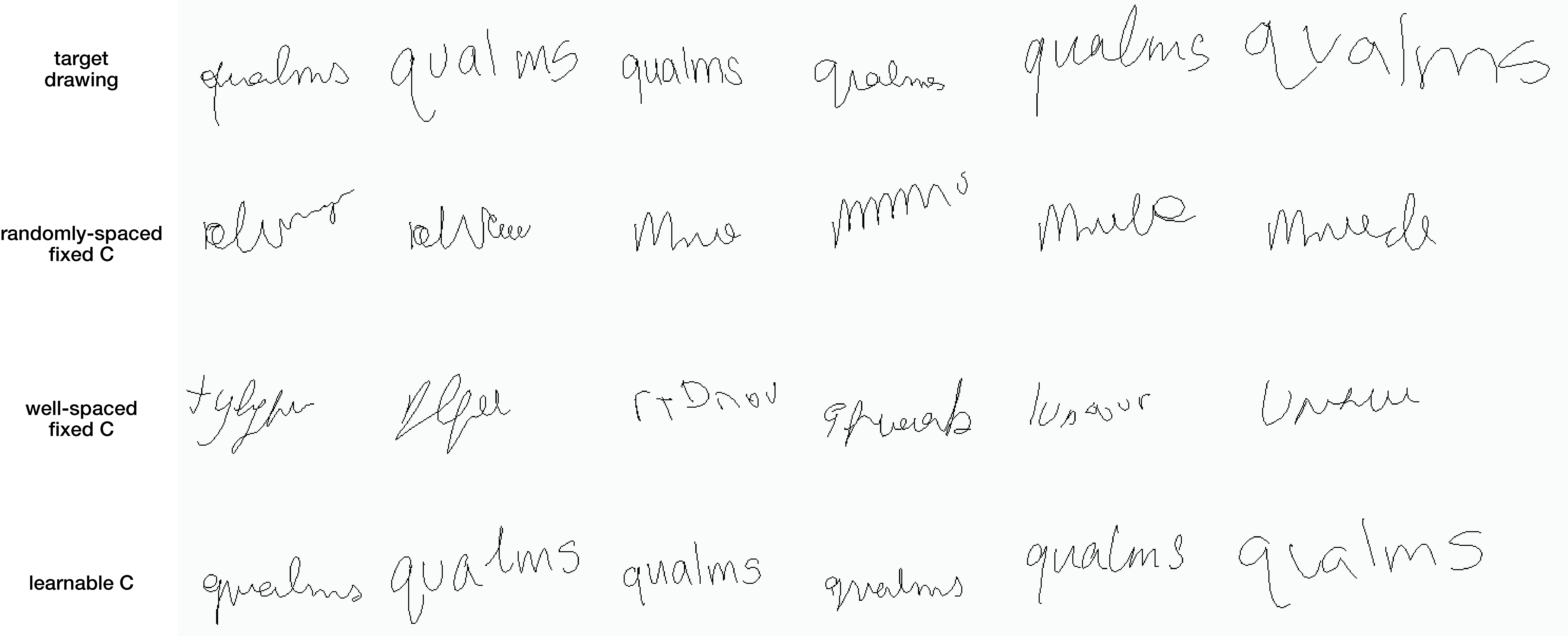}
\caption{Qualitative comparisons of results from different two-character $\textbf{C}$. When $\textbf{C}$ are fixed through training, the models failed to synthesize recognizable letters.}
\label{fig:c_res}
\end{figure}

Figure \ref{fig:c_res} shows writing generation results from these different approaches. Both fixed $\mbC$ approaches fail to generate good samples.

\begin{figure}[t]
\includegraphics[width=1.0\linewidth]{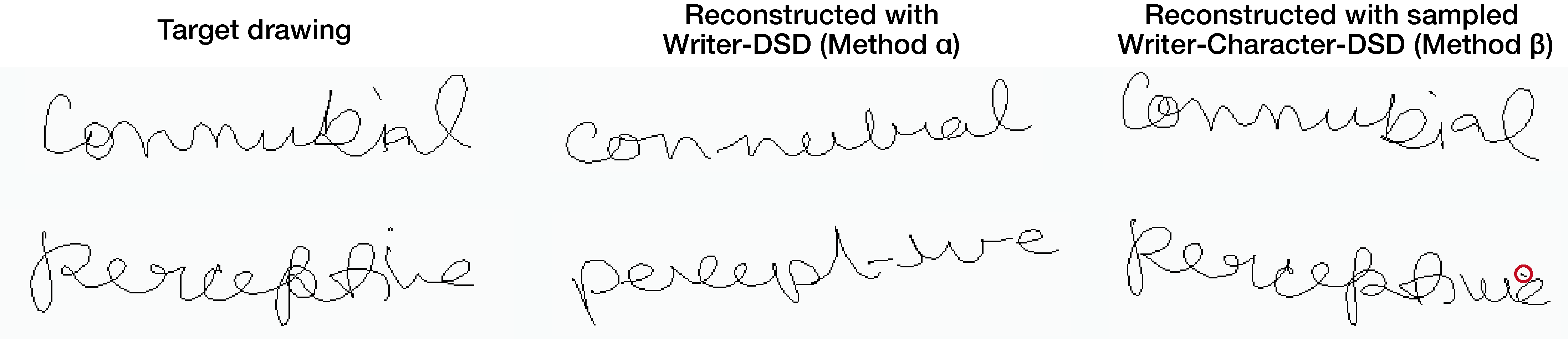}
\caption{Instances of missed delayed strokes by our proposed model.}
\label{fig:missed}
\end{figure}

\paragraph{Limitations of two-character substrings.} One might think that using single- and two-character substrings could represent most variation in writing---how much do letters two behind the currently-written letter really affect the output? Cursive writing especially contains delayed strokes: for example, adding the dot for `i' in `himself' after writing `f'. Changing to two-character substrings removes the ability of our model to learn delayed strokes in writings. In practice, our original $\mbC$ model can struggle to correctly place delayed strokes (Fig.~\ref{fig:missed}): the model must predict a negative x-axis stroke to finish a previous character, which rarely occurs in our training set. This is one area for future work to improve.

\begin{figure}
    \centering
    \includegraphics[width=1.0\textwidth]{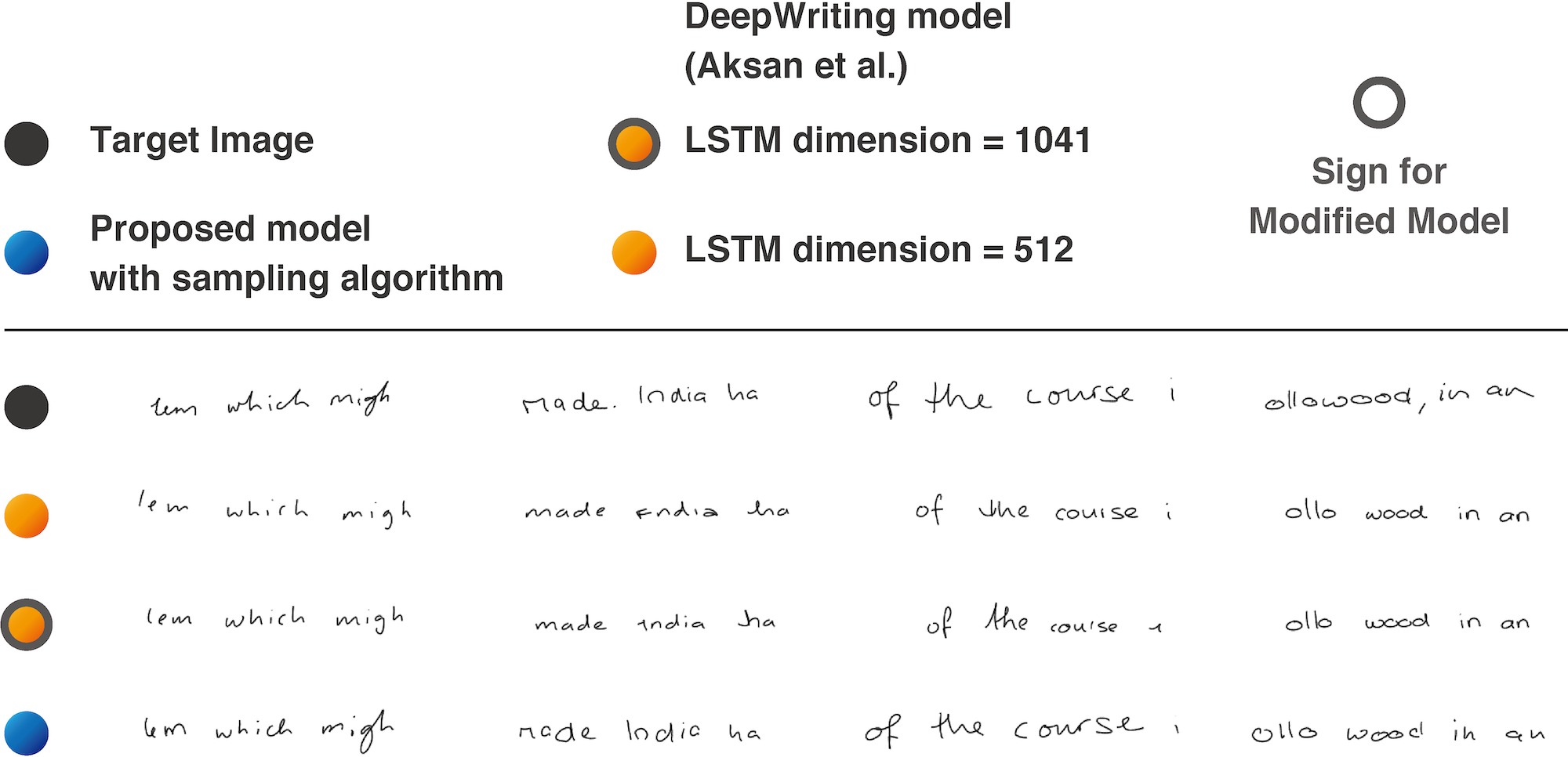}
    \caption{To match the total number of parameters between DeepWriting and our model, we increased the LSTM dimension in DeepWriting from $512$ to $1,041$. There is little improvement in quality from the initial DeepWriting model of $512$ LSTM dimension to $1,041$. These drawings are generated with $10$ sentence-level reference samples of the same writer.}
    \label{fig:lstm_comp}
\end{figure}

\section{Network Capacity}

To validate our network capacity, we conducted two comparison studies with the DeepWriting model by Aksan et al.~\cite{Aksan:2018:DeepWriting}. The first is to decrease the number of parameters in our DSD-based model, and the second is to increase the number of parameters in the DeepWriting model.

\subsection{Increasing DeepWriting Model Parameters}
We modified the hidden state dimension for the DeepWriting model from $512$ to $1,071$, and the total number of parameters subsequently increased from $7.2M$ to $31.3M$. We show a side-by-side comparison of generated samples with DSD-256 model in Figure \ref{fig:lstm_comp}. 
For our 8GB VRAM GPU to accommodate this large LSTM, we decreased the batch size by half from $64$ to $32$, and doubled the number of learning rate decay steps from $1,000$ to $2,000$. However, increasing the capacity of the DeepWriting model did not improve the generated results (Figure \ref{fig:lstm_comp}).

\begin{figure}[t]
    \centering
    \includegraphics[width=0.90\textwidth]{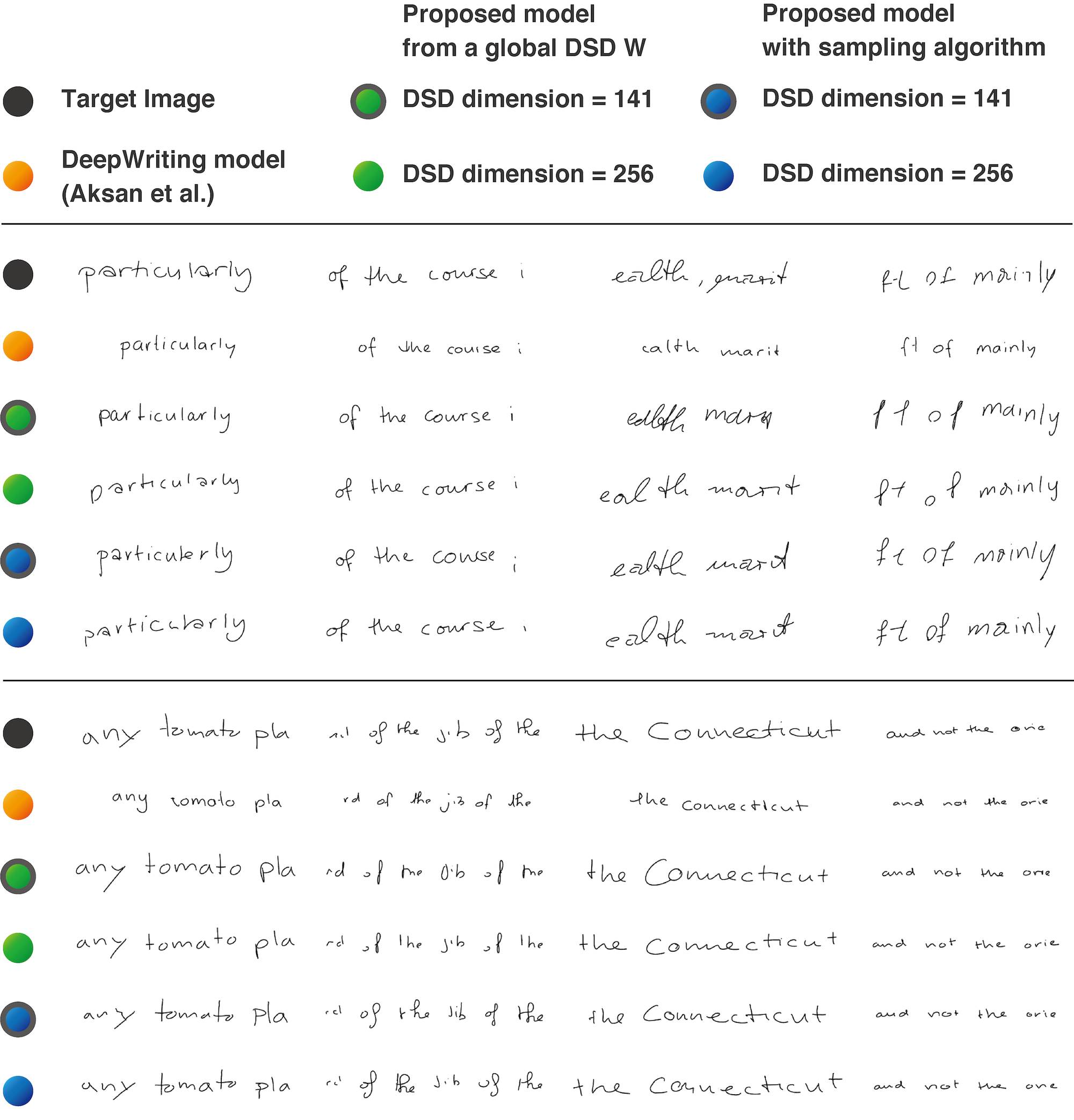}
    \caption{We decreased the DSD dimension in our model from $256$ to $141$ to match the total number of parameters to DeepWriting. As we decrease DSD dimensions, there is a slight fall in quality, particularly the examples with green dots that are generated from a single global writer-DSD $\overline{\mathbf{w}}$ in Method $\alpha$.}
    \label{fig:dsd_comp}
\end{figure}

\subsection{Decreasing DSD Model Parameters}

Is our decoupled model more efficient than the DeepWriting model~\cite{Aksan:2018:DeepWriting}, or simply more capacitive? With $256$-dimensional latent vectors, our model has $31.33M$ parameters, whereas DeepWriting has $7.27M$. 
This difference is largely in the $g_\theta$ fully connected layer which expands $\mbc^{\text{raw}}_{c_t}$ into $\mbC_{c_t}$ via $16.84M$ parameters.
As such, we reduced our latent vector dimension from $256$ to $141$, which leads to a model with $7.25M$ parameters. 
While we observe minor deterioration in generation quality (Figure \ref{fig:dsd_comp}), the model still creates higher-quality samples than DeepWriting. This suggests that our architecture is more efficient.

\section{Further Generated Comparisons}
\begin{figure}[p]
\centering
\includegraphics[width=1.0\textwidth]{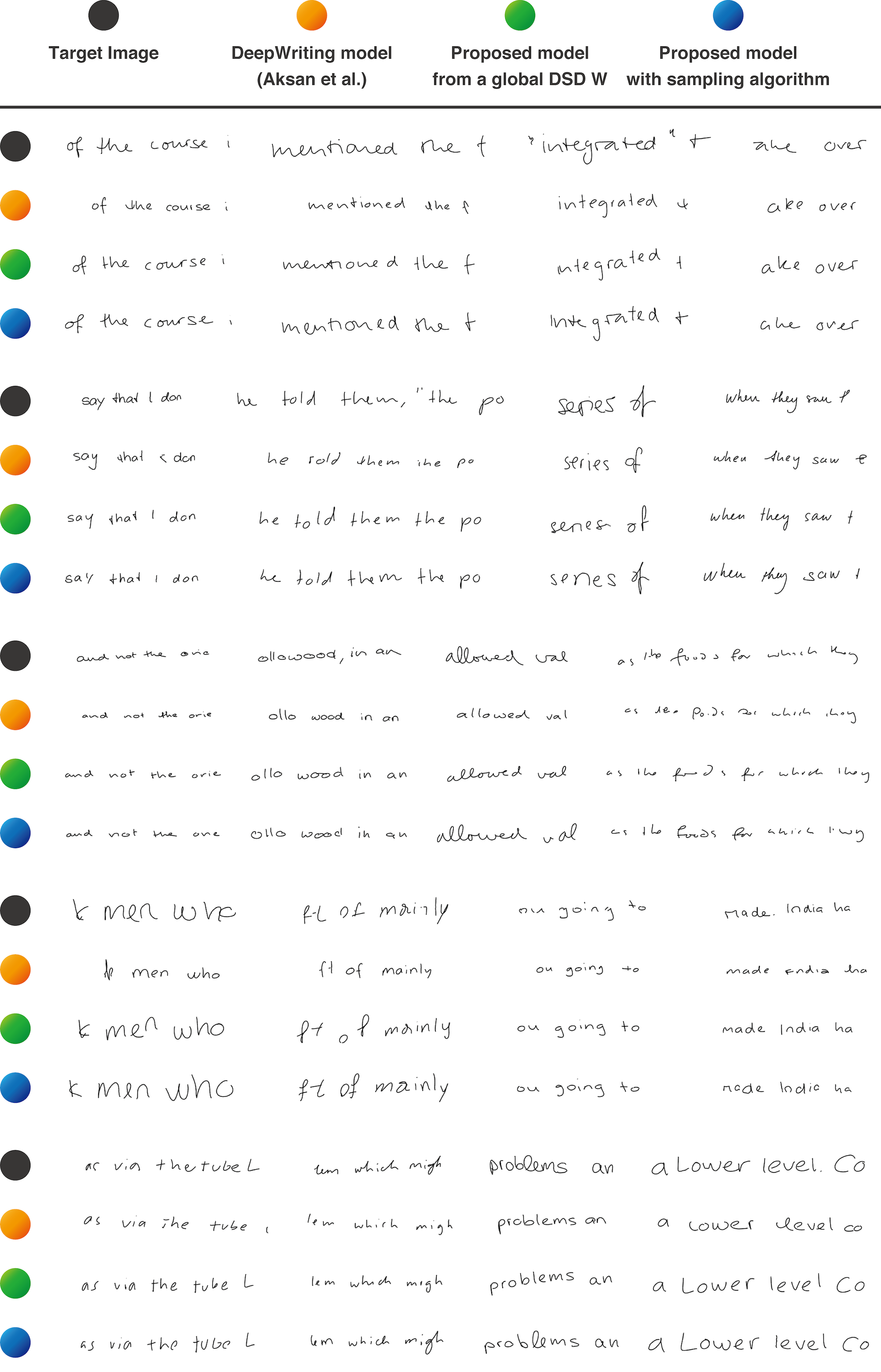}
\caption{The first 20 out of 40 samples used for quantitative evaluation.}
\label{fig:all1}
\end{figure}

\begin{figure}[p]
\centering
\includegraphics[width=1.0\textwidth]{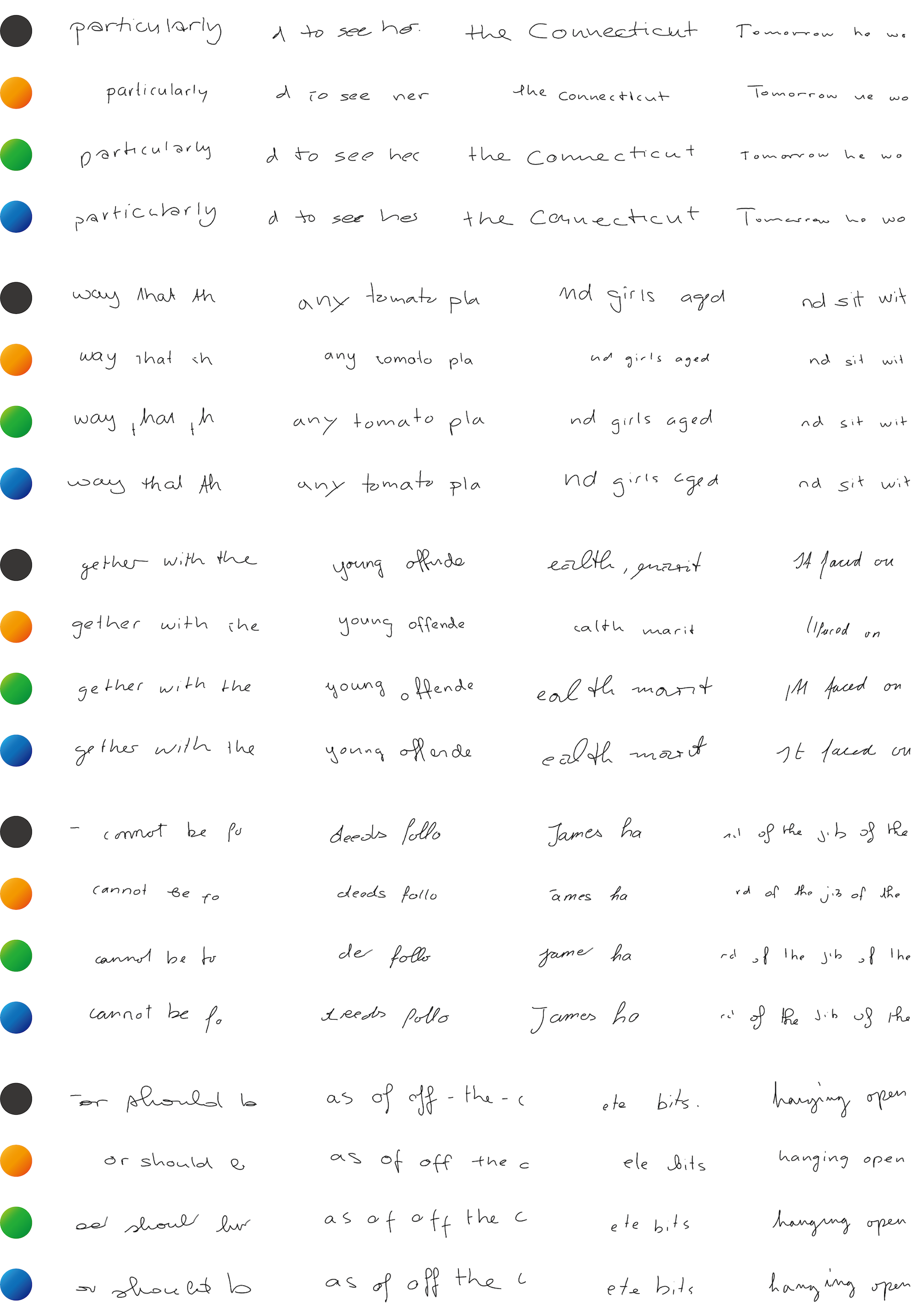}
\caption{The second 20 out of 40 samples used for quantitative evaluation.}
\label{fig:all2}
\end{figure}

Figure \ref{fig:all1} and \ref{fig:all2} show all $40$ samples of drawing used for our qualitative/quantitative study on Amazon Mechanical Turk. 

\clearpage

\section{Sampling Algorithm for Writer-Character-DSD $\mbw_{c_t}$}

When handwriting samples $\mbx$ with corresponding character strings $s$ are provided for inference, we can extract writer-character-DSDs $\mathbf{w}_{c_t}$ from $\mbx$ for substrings of $s$. For example, for character string \textit{his}, we can first extract the following $3$ arrays of writer-character-DSDs using $f_{\theta}^{enc}$:
$[\mbw_{h}]$, $[\mbw_{h}, \mbw_{hi}]$, and $[\mbw_{h}, \mbw_{hi}, \mbw_{his}]$. In addition, if the handwriting is non-cursive and each character is properly segmented, then we can also obtain $3$ more ($[\mbw_{i}]$, $[\mbw_{i}, \mbw_{is}]$, and $[\mbw_{s}]$).
However, we must ensure that the handwriting is cursive, as \textit{h}, \textit{i}, and \textit{s} could be connected by a single stroke. In such cases, we only extract the first $3$ arrays.

We create a database $D$ of these arrays of writer-character-DSDs with substrings as their keys, and query substrings in the target sentence $s^*$ for generation to obtain relevant writer-character-DSDs. We also compute the mean global writer-DSD $\overline{\mbw}$ as $\overline{\mbw} = \frac{1}{N}\sum_{c_t} \mbC_{c_t}^{-1} \mbw_{c_t}$ where $N$ is the number of obtained $\mbw_{c_t}$.

\begin{figure}[t]
\centering
\includegraphics[width=1.0\linewidth]{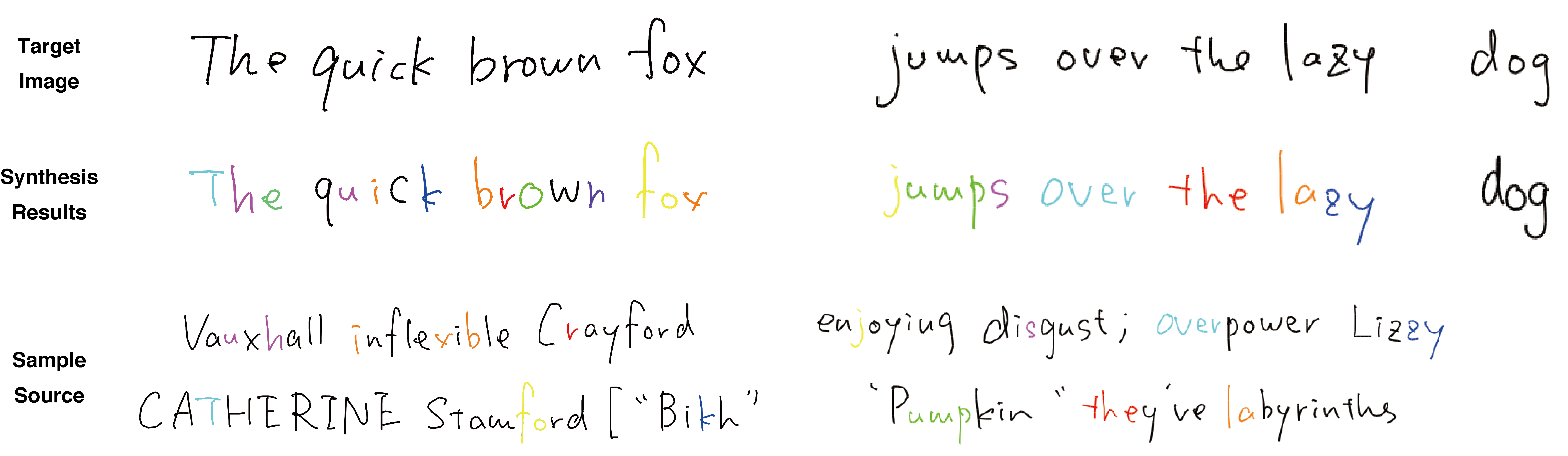}
\caption{Generated image by our sampling algorithm. The black letters in the synthesis indicate that they are predicted from $\overline{\mbw}$, while the colored characters in reference samples are encoded and saved in the database in the form of writer-character-DSDs $\mbw_{c_t}$ and retrieved during synthesis.}
\label{fig:samples}
\end{figure}

To synthesize a sample $thin$ from $his$, we query the substring $hi$ and receive an array of DSDs: $[\mbw_{h}, \mbw_{hi}]$. As $\mbw_{t}$ and $\mbw_{n}$ are computed from $\overline{\mbw}$:
\begin{subequations}\begin{align}
    \mbw_{t}^{\text{rec}} &= h_{\theta}([\mbw_{t}]) \label{aeq:1a} \\
    \mbw_{th}^{\text{rec}} &= h_{\theta}([\mbw_{t},\mbw_{h}]) \label{aeq:1b} \\
    \mbw_{thi}^{\text{rec}} &= h_{\theta}([\mbw_{t},\mbw_{hi}]) \label{aeq:1c} \\
    \mbw_{thin}^{\text{rec}} &= h_{\theta}([\mbw_{t},\mbw_{hi},\mbw_{n}]) \label{aeq:1d}
\end{align}\end{subequations}
We use $[\mbw_{t},\mbw_{hi}]$ instead of $[\mbw_{t},\mbw_{h},\mbw_{hi}]$ in Equations \ref{aeq:1c} and \ref{aeq:1d} because, as one might recall from generation Method $\beta$ in the main paper (Sec.~3), the function approximator $h_{\theta}$ is designed to \textit{restore} temporal dependencies between writer-character-DSDs. As `h' and `i' are already temporally \textit{dependent} within $\mbw_{hi}$, we need only connect characters `t' and `h' through LSTM $h_{\theta}$. The pseudocode for this sampling procedure is shown in Algorithm \ref{alg:1}, with example generations in Figure \ref{fig:samples}.

\SetArgSty{textnormal}
\begin{algorithm}[t]
\caption{Pseudocode for our sampling algorithm to reconstruct writer-character-DSDs for the target sentence to synthesize.}
\label{alg:1}
\SetAlgoLined
\DontPrintSemicolon
\KwInput{$D$: database of writer-character-DSD, $s^*$: target sentence to generate, $\overline{\mbw}$: mean global writer-DSD\;}
\SetKwFunction{FMain}{PerformSamplingAlgorithm}
\SetKwProg{Fn}{Function}{:}{}
\Fn{\FMain{$D$, $s^*$, $\overline{\mbw}$}}{
	Initialize empty sets $L$, $R$ and $result$\;
	$s^*$ $\leftarrow$ MarkAllCharactersAsUncovered($s^*$)\;
	$\mathbf{ss}^*$ $\leftarrow$ ExtractSubStringsAndOrderByLength($s^*$)\;
	\For{each substring $ss$ in $\mathbf{ss}^*$}{
    	\If{$ss \text{ is in } D \text{ and every characters in } ss \text{ are not-covered}$}{
        	$[\mbw_{c_1}, ..., \mbw_{c_t}]$ = QueryDatabaseWithKey($ss$)\;
            Add $[\mbw_{c_1}, ..., \mbw_{c_t}]$ to $L$\;
            $\mbs^*$ $\leftarrow$ MarkCharactersInSubstringAsCovered($\mbs^*$, $ss$)
        }
    }
	\For{each uncovered character $c_t$ in $s^*$}{
    	$\mbw_{c_t} \leftarrow \mbC_{c_t} \overline{\mbw}$\;
        Add $[\mbw_{c_t}]$ to $L$\;
    }
	$L^*$ $\leftarrow$ OrderSetBySubstringAppearanceIn($s^*$)\;
	\For{each array $A$ in $L^*$}{
    	\For{each $\mbw_{c_i}$ in $A$}{
			$\mbw_{c_i}^{\text{rec}} \leftarrow h_{\theta}([R_1, R_2, ..., \mbw_{c_i}])$\;
            Add $\mbw_{c_i}^{\text{rec}}$ to the $result$ list\;
			\If{$\mbw_{c_t}$ is the last element in $A$}{
            	Add $\mbw_{c_i}$ to the reference set $R$\;
            }
        }
    }
    \Return{$result$}
}
\end{algorithm}

\section{Sequence Decoder $f_{\theta}^{\text{dec}}$}

\begin{figure}[t]
\centering
\includegraphics[width=1.0\linewidth]{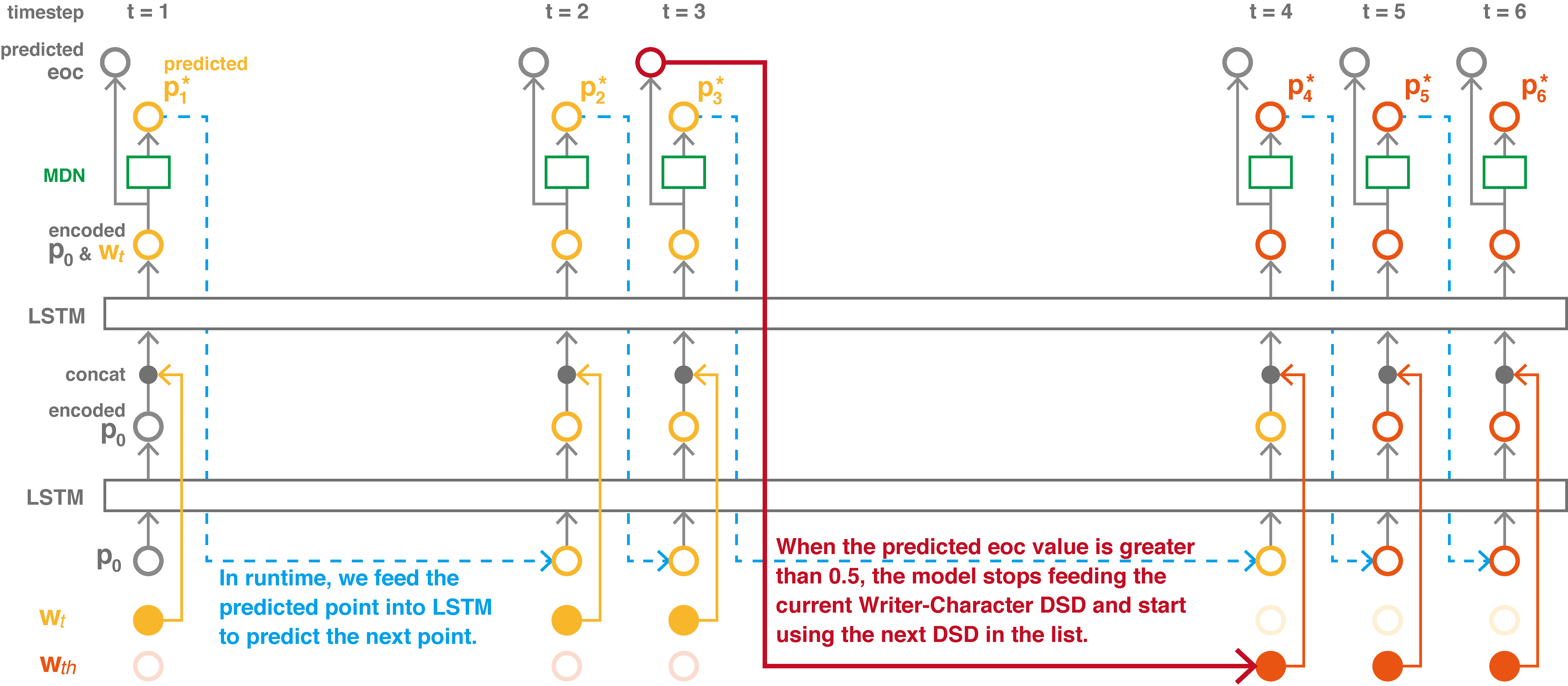}
\caption{Overview of our decoder architecture. During training, we feed true point sequences to the LSTM and do not use the predicted output $p_t^*$ as the next input (the procedure shown as dotted blue lines).}
\label{fig:decoder}
\end{figure}

To synthesize a new sample from a list of writer-character-DSD $\mbw_{c_t}$, we train a sequence decoder function $f_{\theta}^{\text{dec}}$. The inputs to this decoder are: 1) initial point $p_0 = (0,0,0)$, and 2) the first writer-character-DSD  $\mbw_{c_1}$. Continuing with the \textit{thin} example, we predict the first point $p_1$ from $p_0$ and $\mbw_{t}$. At runtime, the predicted point $p_1^*$ will be fed into the LSTM at the next timestep to predict $p_2$. When the decoder model outputs an $eoc > 0.5$ (end-of-character probability), the model stops drawing the current character and start referencing the next writer-character-DSD so that it starts drawing the next character. This procedure is illustrated as the red lines in Figure \ref{fig:decoder}. Similarly, to determine the touch/untouch status of the pen to the canvas, we use the $eos$ (end-of-stroke probability) which is enclosed in point prediction $p_t^*$. If $eos_t > 0.5$, our model lifts up the pen; if $eos_t\leq 0.5$, our model continues the stroke. 

Note that when we use the predicted $p_t^*$ as an input to the LSTM at runtime, we binarize the $eos$ value. This is because all $eos$ values in training data are binarized. Further, we do not use the predicted points to predict the next point during training, because we have the true point sequence $\mbx$. In other words:
\begin{subequations}
\begin{align}
	p_{t+1}^*  &= f_{\theta}^{\text{dec}}(p_{0}, p_{1}, ..., p_{t} | \mbw_{c_t})\qquad (training) \\
	p_{t+1}^*  &= f_{\theta}^{\text{dec}}(p_{0}, p_{1}^*, ..., p_{t}^* | \mbw_{c_t})\qquad	(runtime)
\end{align}
\end{subequations}
where $*$ indicates \textit{predicted} outputs by the decoder network.

\begin{figure}[t]
\centering
\includegraphics[width=0.9\linewidth]{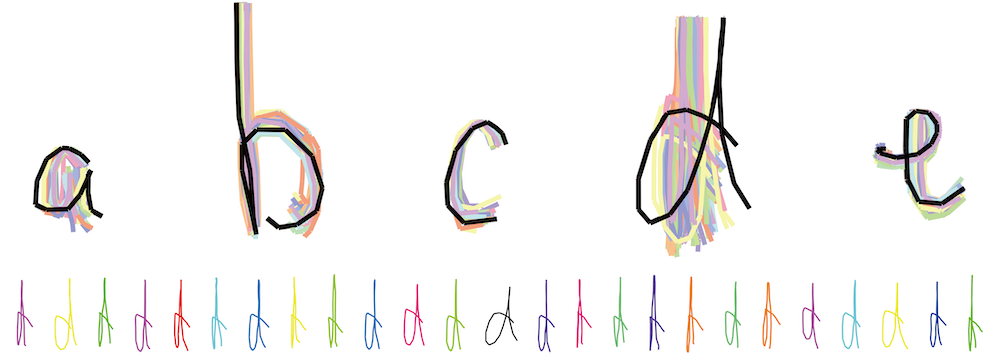}
\caption{Variations in generated results from a single writer-character-DSD $\mbw_{c_t}$, achieved by sampling points from predicted MDN distributions.}
\label{fig:intra}
\end{figure}

Finally, the mixture density networks \cite{bishop1994mixture} (MDN) layer in our decoder makes it possible for our model to generate varying samples even from the same writer-character-DSD $\mbw_{c_t}$. Examples are shown in Figure \ref{fig:intra}.

\section{Character Encoder Function $g_\theta$}

Next, we discuss in detail how the character matrix $\mbC$ is computed. First, we convert each one-hot character vector $\mbc_t$ in the sentence $\mbs$ into a $256$ dimensional vector $\mbc_t^{\text{raw}}$ via a fully-connected layer $g_{\theta}^{\text{FC1}}$. Then, we feed this vector into LSTM $g_\theta^{\text{LSTM}}$ and receive outputs $\mbc_{c_t}^{\text{raw}}$ of the same size. $g_\theta^{\text{LSTM}}$ is designed to encode temporal dependencies among characters. Then, we use a mapping function $g_{\theta}^{\text{FC2}}$ to transform the $256\times1$ vector into a $65,536$ dimensional vector, and finally reshape the output vector to a $256\times 256$ matrix $\mbC_{c_t}$. This process is as follows:
\begin{subequations}
	\begin{align}
		\mbc_t^{\text{raw}} &= g_{\theta}^{\text{FC1}}(c_t)\label{eq:7a}\\
		\mbc_{c_t}^{\text{raw}} &= g_{\theta}^{\text{LSTM}}([\mbc_1^{\text{raw}}, ..., \mbc_t^{\text{raw}}])\label{eq:7b}\\
		\mbC_{c_t} &= Reshape(g_{\theta}^{\text{FC2}}(\mbc_{c_t}^{\text{raw}}))\label{eq:7c}
	\end{align}
\end{subequations}

The parameters in $g_{\theta}^{\text{FC2}}$ take up about one third of total number of parameters in our proposed model; this is expensive.
However, using a fully-connected layer allows each value in the output $\mbC_{c_t}$ to be computed from all values in the 256-dimensional vector $\mbc^{\text{raw}}_{c_t}$. 
If each value in $\mbc_{c_t}^{\text{raw}}$ represents some different information about the character, then we intended to weight them $65,536$ times via distinct mapping functions to create a matrix $\mbC_{c_t}$. 
We leave the study of other possible $g_{\theta}$ architectures for future work.

\section{Segmentation Network $k_{\theta}$}
\label{sec:segmentationnetwork}
\label{sec:segnet}

We introduce an unsupervised training technique to segment sequential handwriting samples into characters without any human intervention. For comparison, the existing state-of-the-art DeepWriting handwriting synthesis model \cite{Aksan:2018:DeepWriting} relies on commercial software for character segmentation.

Our data samples for training arrive as stroke sequences and character strings, with no explicit labeling on where one character ends and another begins within the stroke sequence. As such, we train a segmentation network $k_{\theta}$ to segment sequential input data $\mbx$ into characters, and to predict \textit{end of character (eoc)} labels for each point in $\mbx$. Relying on these predicted \textit{eoc} labels, we can extract $\mbw_{c_t}$ from encoded $\mbx^*$ and synthesize new samples with $f_{\theta}^{\text{dec}}$.

\begin{figure}[t]
	\centering
	\includegraphics[width=1.0\linewidth]{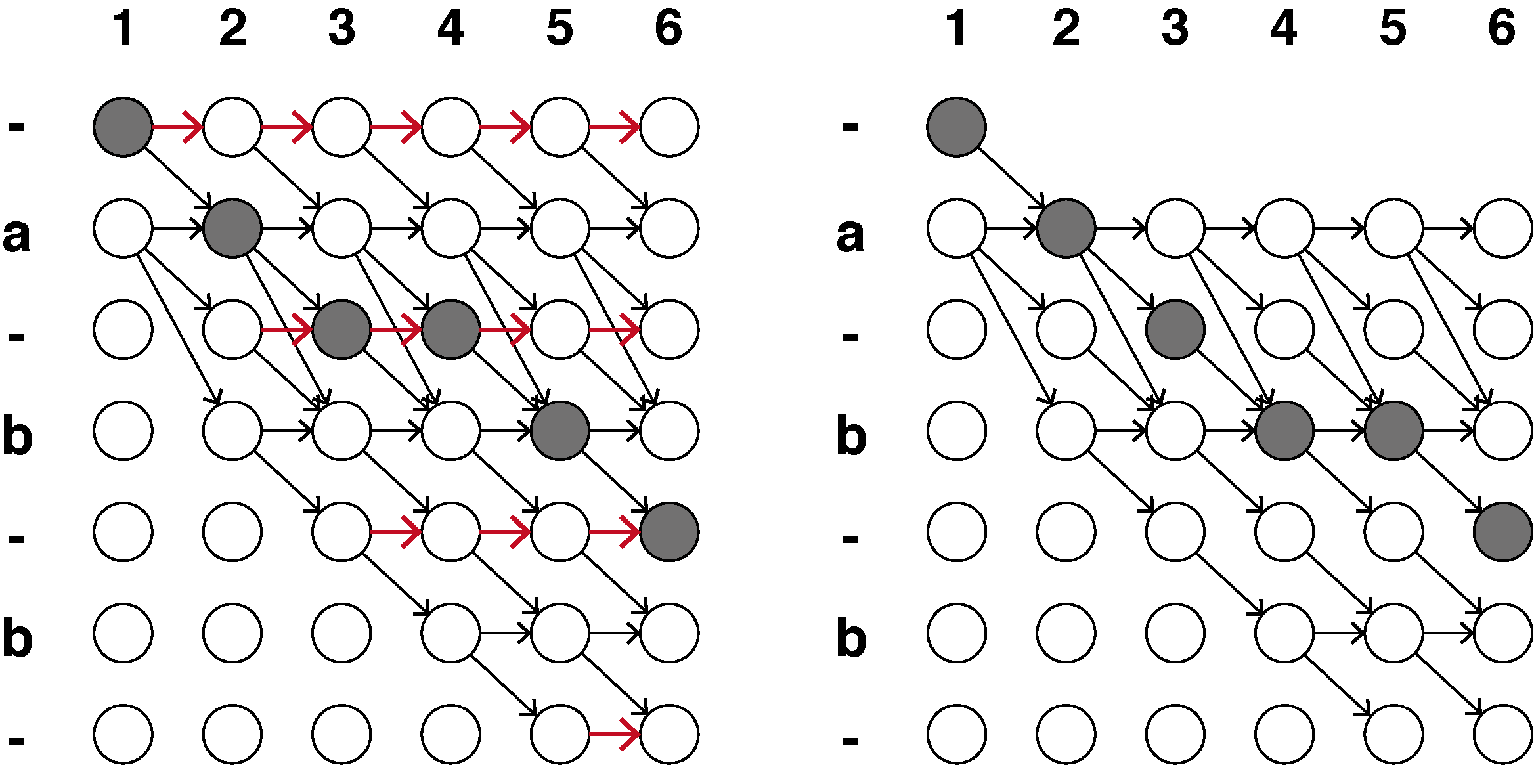}
    \caption{Illustration of connections between temporal nodes. (Left) Original CTC connections. (Right) Our design of CTC connections. The connections between non-character nodes `-' are prohibited (red arrows in the original). The shaded nodes shows an example route for prediction.}
    \label{fig:ctc}
\end{figure}

\begin{figure}[t]
	\centering
	\includegraphics[width=1.0\linewidth]{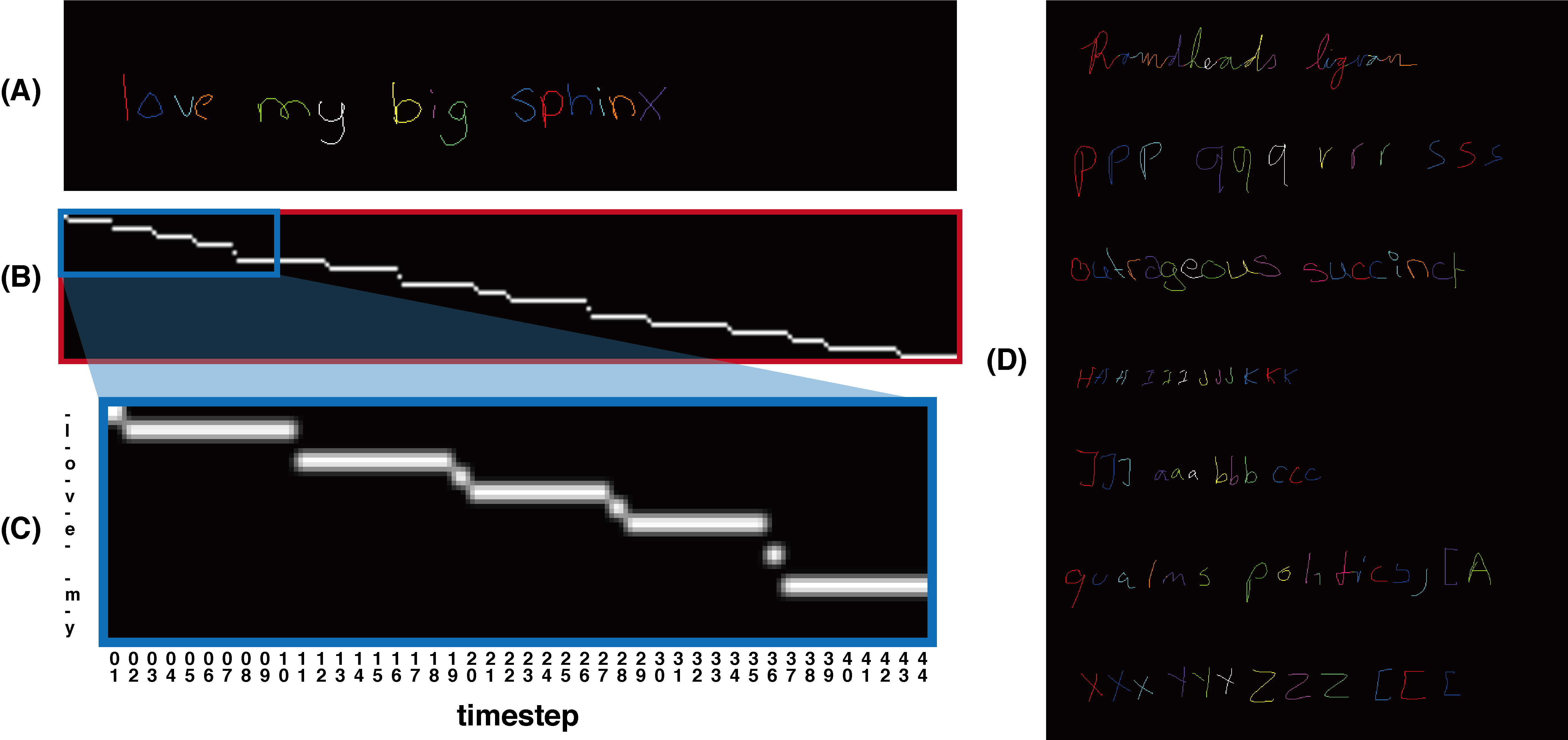}
    \caption{Segmentation results. A) in handwriting image format. Different colors indicate different character segments. B) in CTC best route format. C) enlarged figure of the route path. D) More results.}
    \label{fig:segseg}
\end{figure}

To prepare the input data for training, we extract $23$ features per point $p_t$ in $\mbx$, as is commonly used in previous work \cite{graves2008novel,NPen-jaeger2001online,keysers2016multi}. We feed these into a bidirectional LSTM to output a probability distribution over all character classes. From this output $O$ of size $(N, Q)$, where $N$ is the input sequence length and $Q$ is the total number of characters, we compute a loss that is similar to a connectionist temporal classification (CTC) loss \cite{ctc}. As seen in Figure \ref{fig:ctc}, we make a slight modification in connections among nodes to adjust the change in two domains: character recognition and segmentation. In the recognition task, the blank character \textit{-} was introduced to fill the gap between two character predictions (e.g., \textit{a--b--b}), but because our goal is to label each point in the input sequence with a specific character in the corresponding sentence, we must avoid unnecessary use of the blank character and instead predict actual characters (e.g., \textit{aaabbbb}). The only case where the blank character is needed in segmentation is when a character is repeated in a sentence. To highlight the switch from the first \textit{b} case to the second \textit{b} case, we use the \textit{-} (e.g., \textit{aaabb-b}). This slight modification in CTC connections enables us to train our segmentation network in an unsupervised manner, automatically label sequential handwriting data with characters and identify \textit{eoc} indices. Examples of segmentation are shown in Figure \ref{fig:segseg}.

\section{Detailed Training Procedure}

\subsection{Ablation Study}
\begin{figure}[t]
    \centering
    \includegraphics[width=1.0\linewidth]{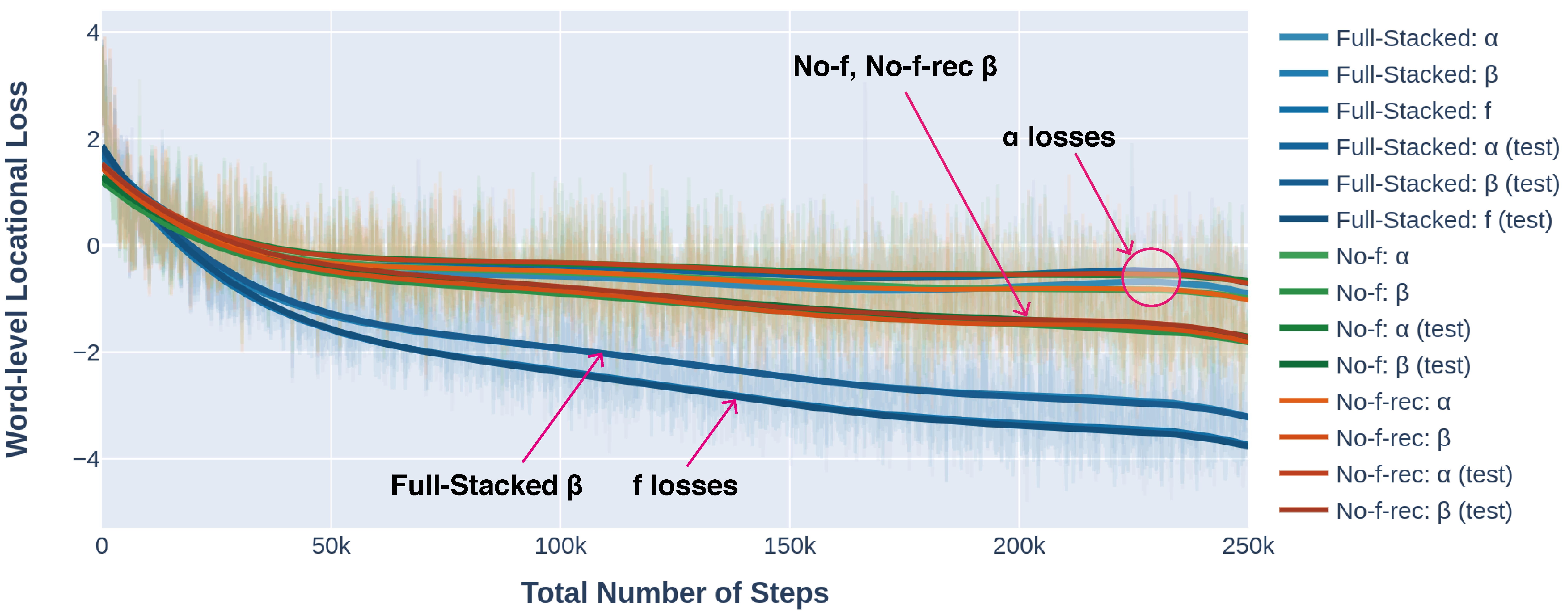}
    \caption{Effects of $\mathscr{L}_{f_{\theta}^{\text{enc}}}$ on training word-level location loss $\mathscr{L}^{word}_{loc}$. Transparent lines show the actual data points, and solid lines show smoothed training curves. \textit{Full-Stacked model} is trained with the full loss, while \textit{No-f} model's loss function does not include $\mathscr{L}_{f_{\theta}^{\text{enc}}}$. Further, \textit{No-f-rec} model does not have $\mathscr{L}_{f_{\theta}^{\text{enc}}}$, $\mathscr{L}_{\alpha}^{\mbw_{c_t}}$, $\mathscr{L}_{\beta}^{\mbw_{c_t}}$ terms in its loss function. \textit{Test} data is $20$ held-out writers.}
    \label{fig:train_plot}
\end{figure}

In the main paper, we discussed three different ways to obtain $\mbw_{c_t}$: $f_{\theta}^{\text{enc}}$, Method $\alpha$, and Method $\beta$. As we compute losses for each type, we conducted a simple ablation study. First, removing $\mathscr{L}_{\alpha}$ from the total loss will take away the ability to generate handwriting samples from the mean writer-DSD $\overline{\mbw}$ from the model by construction. Similarly, excluding $\mathscr{L}_{\beta}$ will disallow our model to synthesize a new sample from saved writer-character-DSDs in the database $D$ and only allow it to generate from the mean $\overline{\mbw}$. It is clear that we need both types of losses, $\mathscr{L}_{\alpha}$ and $\mathscr{L}_{\beta}$, to have the current model capabilities. 

However, eliminating $\mathscr{L}_{f_{\theta}^{\text{enc}}}$, the loss term for a method that uses the original writer-character-DSD extracted from $f_{\theta}^{\text{enc}}$, does not change model dynamics. Hence, we trained ablated models with modified loss functions that do not include: 1) any terms related to $f_{\theta}^{\text{enc}}$ (i.e., $\mathscr{L}_{f_{\theta}^{\text{enc}}}$, $\mathscr{L}_{\alpha}^{\mbw_{c_t}}$ and $\mathscr{L}_{\beta}^{\mbw_{c_t}}$), and 2) just $\mathscr{L}_{f_{\theta}^{\text{enc}}}$. Figure \ref{fig:train_plot} shows the training curve for word-level location losses $\mathscr{L}^{loc}$. Removing $\mathscr{L}_{f_{\theta}^{enc}}$ had significant influence on Method $\beta$ locational loss, $\mathscr{L}^{loc}_{\beta}$ (standard: $-3.213$ vs. ablated: $-1.811$ after 250K training steps). 

From this result, we assume that $\mathscr{L}_{f_{\theta}^{\text{enc}}}$ works as a learning guideline for our model, and speeds up the training. We analyze that this is because having $\mathscr{L}_{f_{\theta}^{\text{enc}}}$ in our loss function encourages accurate learning for our decoder function $f_{\theta}^{\text{dec}}$. In this setting, the function $f_{\theta}$ is indeed an autoencoder, and the decoder is trained to restore $\mbx$ from its encoded representation, writer-character-DSDs.
This will increase the decoder performance, and as the decoder accuracy is maintained, the model can focus on learning the encoder problem, which is reconstruction of writer-character-DSDs by Method $\alpha$ and $\beta$.

The reconstruction losses, $\mathscr{L}_{\alpha}^{\mbw_{c_t}}$ and $\mathscr{L}_{\beta}^{\mbw_{c_t}}$, by contrast, did not affect the learning speed. We assume this can also be addressed by the same reason. 
Even if we constrained the reconstructed DSDs by Method $\alpha$ and $\beta$ to minimize their differences with the original DSDs from $f_{\theta}^{\text{enc}}$, those constraints will penalize the encoder more than they do for the decoder. To effectively train the decoder function, our model thus requires the loss term $\mathscr{L}_{f_{\theta}^{\text{enc}}}$.

\subsection{Hyperparameters}

To train our synthesis model, we use Adam~\cite{adam14} as our optimizer and set the learning rate to $0.001$. 
We also clip the gradients in the range $[-10.0, 10.0]$ to enhance learning stability. 
We use $5$ sentence-level samples (relevant word-level and character-level samples are included as well) for each batch in training. We use multi-stacked ($3$-layers) LSTMs for our recurrent layers.

\section{Dataset Specification and Collection Methodology}
\label{sec:dataset}

\begin{figure}[t]
    \centering
    \includegraphics[width=1.0\linewidth]{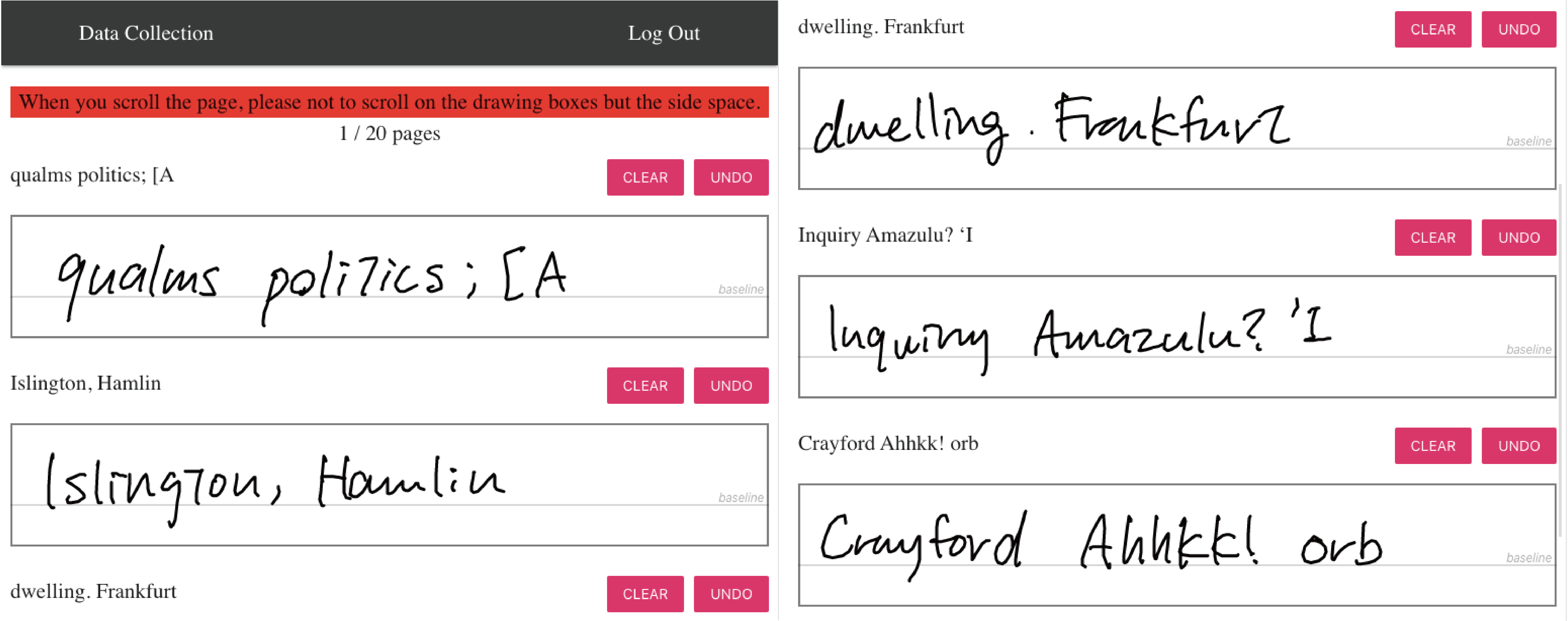}
    \caption{Example screen of our data collection website. Each drawing box is $750$ pixels $\times$ $120$ pixels, and we provide a baseline at $80$ pixels from the top.}
    \label{fig:web}
\end{figure}

Our dataset considers the $86$ characters: a space character ` ', and the following $85$ characters:
\begin{verbatim}
0123456789
abcdefghijklmnopqrstuvwxyz
ABCDEFGHIJKLMNOPQRSTUVWXYZ
!?"'*+-=:;,.<>\/[]()#$%&
\end{verbatim}

We collected handwriting samples from $170$ writers using Amazon Mechanical Turk. An example screen of our data collection website is shown in Figure \ref{fig:web}. Writing arbitrary words is laborious, and so we set a data-collection time limit of $60$ minutes. Given this, it was necessary to select a subset of English words for our data collection. 

\subsection{Defining Target Words and Sentences}

We analyzed the Gutenberg Dataset \cite{lahiri:2014:SRW}, which is a large corpus of $3,036$ English books. These documents use $99$ characters in total, including alphabetical, numerical, and special characters. In total, $5,831$ character pairs appear in the dataset, while theoretically there are $9,702$ possible character pairs ($99\times 99 - 99$). By counting the number of occurrences of each character pair, we constructed an ordered list of character pairs that is then used to score $2,158,445$ distinctive words within the corpus. 

The first word to be selected from the corpus was \textit{therefore}, which includes the two most frequently used character pairs \textit{th} and \textit{he}. In fact, the character pairs within \textit{therefore} appear so frequently that they altogether cover $13.5\%$ of all character pair occurrences. 

After adding \textit{therefore} to the list of words for experiments, we then add additional words iteratively: we re-calculate scores for all other words with updated scores of character pairs (i.e., after adding \emph{therefore}, the pairs \textit{th} and \textit{he} will not have high scores in future iterations). This process was repeated until the words in the list exceeded $99\%$ coverage of all character pair occurrences. 

Then, we constructed sentences from these high-scored words. Each sentence was less than 24 characters length to meet a space constraint due to our experiment setup. We asked tablet owners to write the prompted sentences using their stylus within the bounding box ($750$ pixels $\times$ $120$ pixels), and $24$ characters was the maximum number of characters that could reasonably fit into the region. 

We also added several pangram sentences as well as repeated characters sentences (e.g., \textit{aaa bbb ccc ddd}), and that led to our basic list of $192$ sentences. These sentences use $86$ unique characters, instead of $99$ available characters, due to our decision to ignore rarely used special characters. 
They also use $1,182$ distinct character pairs which cover $99.5\%$ of all character pair occurrences ($1,158,051,103/1,164,429,941$). The remaining pairs could have been ignored, yet because that $0.5\%$ was still large---$6,378,838$ occurrences by $4,649$ character pairs---we decided to create a list of extra words with less frequently used character pairs, distribute them to $170$ writers. Thus, each writer creates some rarer data that varies for each writer, in addition to their basic $192$ sentences. As a result, we achieve $99.9\%$ coverage with $3,894$ character pairs.

\subsection{Writer Behavior}

Handwriting dataset collection is complex for various reasons, and in general creating a clean dataset without heuristic or manual cleaning is difficult. In our collection process, sometimes a writer would realizes that s/he missed certain characters in the sentence after finishing the line, and so would go back to the location to add new strokes. These `late' character additions are accidental rather than intentional. In contrast, conventional online handwriting recognition literature defines \textit{delayed strokes}, where in cursive writing the horizontal bar of \textit{t} and \textit{f}, or the dot of \textit{i} and \textit{j}, are often added after a writer finished the current word. To distinguish between these two cases of late characters and delayed strokes, we disregard the temporal order of each stroke in a sample and reorder them from left to right \emph{if} the leftmost point in a stroke is to the right of the rightmost point in another stroke. In this way, accidental omissions are removed.

Further, although we strongly advised participants to erase previous lines if they made mistakes, most participants either ignored this and left mistakes in, or scribbled over those regions to block them out. Writers also missed characters from the prompted sentences, and not a single participant (out of $170$ writers) succeeded in near-perfect writing of $192$ sentences. 
As our segmentation network (Sec.~\ref{sec:segmentationnetwork}) assumes that each drawing sample is labeled with the accurate character sequence, missed characters can directly affect the performance of segmentation. Hence, we manually corrected these instances throughout our dataset.

\end{document}